\documentclass[10pt,twocolumn,letterpaper]{article}

\usepackage{cvpr}              

\usepackage[dvipsnames]{xcolor}

\usepackage{url}

\usepackage{adjustbox}
\usepackage{arydshln}
\usepackage[accsupp]{axessibility}
\usepackage{booktabs}
\usepackage{graphicx}
\usepackage{multirow}
\usepackage{subcaption}

\definecolor{cvprblue}{rgb}{0.21,0.49,0.74}
\usepackage[pagebackref,breaklinks,colorlinks,citecolor=cvprblue]{hyperref}

\def\paperID{8422} 
\def\confName{CVPR}
\def\confYear{2024}

\title{Personalized Residuals for Concept-Driven Text-to-Image Generation}

\author{Cusuh Ham\thanks{Work performed during an internship at Adobe Research.}\\
Georgia Institute of Technology\\
{\tt\small cusuh@gatech.edu}
\and
Matthew Fisher\\
Adobe Research\\
{\tt\small matfishe@adobe.com}
\and
James Hays\\
Georgia Institute of Technology\\
{\tt\small hays@gatech.edu}
\and
Nicholas Kolkin\\
Adobe Research\\
{\tt\small kolkin@adobe.com}
\and
Yuchen Liu\\
Adobe Research\\
{\tt\small yuliu@adobe.com}
\and
Richard Zhang\\
Adobe Research\\
{\tt\small rizhang@adobe.com}
\and
Tobias Hinz\\
Adobe Research\\
{\tt\small thinz@adobe.com}
}

\begin{document}
\maketitle
\begin{abstract}
We present \textit{personalized residuals} and \textit{localized attention-guided sampling} for efficient concept-driven generation using text-to-image diffusion models. Our method first represents concepts by freezing the weights of a pretrained text-conditioned diffusion model and learning low-rank residuals for a small subset of the model's layers. The residual-based approach then directly enables application of our proposed sampling technique, which applies the learned residuals only in areas where the concept is localized via cross-attention and applies the original diffusion weights in all other regions. Localized sampling therefore combines the learned identity of the concept with the existing generative prior of the underlying diffusion model. We show that personalized residuals effectively capture the identity of a concept in $\sim$3 minutes on a single GPU without the use of regularization images and with fewer parameters than previous models, and localized sampling allows using the original model as strong prior for large parts of the image.
\end{abstract}    
\section{Introduction}
\label{sec:intro}

Large-scale text-to-image diffusion models have demonstrated the ability to generate high-quality images that follow the constraints of the input text \cite{saharia2022photorealistic,rombach2022high,ramesh2022hierarchical}.
However, these models do not inherently encode any information about the \textit{identity} of a specific concept, thus limiting the control over specifying a particular instance to appear in the generated image.
To address this, recent approaches propose techniques to \textit{personalize} these models such that they can generate specific concepts in novel environments and styles.

Given a set of images depicting the desired concept, personalization approaches differ in which parameters they train and whether they are specific to a single concept (i.e., they need to be separately trained for each new concept) or can generalize to new concepts without retraining.
To enable personalization of arbitrary concepts, one can finetune the model's parameters \cite{ruiz2023dreambooth} or its inputs \cite{gal2022image} directly such that it can reconstruct the training data.
These approaches can be applied to any kind of concepts, but the finetuning needs to be done on a per-concept basis and different parameters need to be stored for each.
Other approaches train an encoder specific to a particular domain (e.g., faces) and finetune the diffusion model once to use the encoder's embeddings to reconstruct specific concepts within that domain \cite{xiao2023fastcomposer,gal2023encoder,ruiz2023hyperdreambooth}.
The advantage of the latter approach is that it does not require retraining for every concept and can instead be used to instantly generate new concepts from the given domain. However, this approach is limited to a single domain and requires a large dataset to train the encoder.

\begin{figure*}[t]
    \centering
    \includegraphics[width=\textwidth]{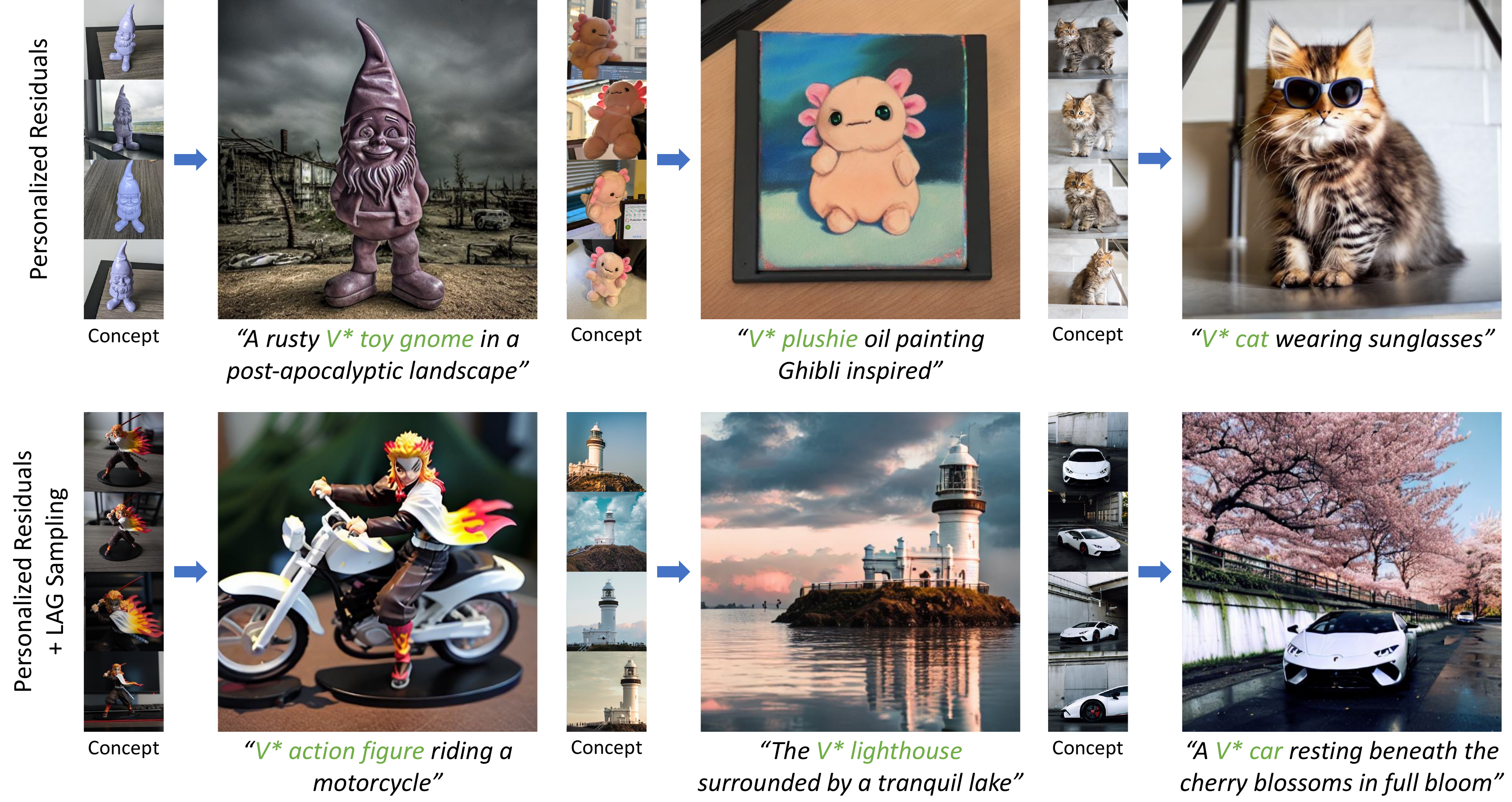}
    \caption{(Top) Given a set of reference images, we learn \textit{personalized residuals} for a subset of a pretrained diffusion model's weights for efficient concept-driven text-to-image generation. (Bottom) The residuals can be combined with our proposed \textit{localized attention-guided (LAG) sampling}, which leverages the cross-attention maps from the diffusion models to localize the application of the residuals and uses the original, unchanged, diffusion model for generating everything else.\vspace{-0.5em}}
    \label{fig:teaser}
\end{figure*}

Our approach follows the former setting, i.e., it finetunes the model's parameters for each concept so that there are no constraints on the domain (see \Cref{fig:teaser} for examples using our proposed method).
The main challenges of open-domain approaches is the need for regularization to mitigate forgetting of concepts learned in the model's original training, and the computational overhead in finetuning a new set of parameters for each concept.
The most common regularization approach is to use images from the same domain as the target concept with the reference images during the finetuning of parameters.
The choice of regularization images affects the quality of the final outputs and, as such, is usually model-, training-, and sometimes even concept-dependent.
Finally, to address the large overhead of finetuning a whole new model for each concept, many approaches only finetune a subset of parameters (e.g., attention layers weights \cite{kumari2023multi}) or the input to the text-to-image model (e.g., the text embedding representing a specific concept \cite{gal2022image}).

Our approach further reduces the number of learnable parameters and does not rely on regularization images.
While most approaches focus on finetuning the key and value weights of the cross-attention layers, we instead predict a low-rank residual \cite{hu2021lora} to the weights of the output projection conv layer after each cross-attention layer.
This allows us to finetune even fewer parameters (about $\sim$0.1\% of the base model) than previous approaches.
Furthermore, we find that this approach does not require any regularization images which makes our approach both simpler, since we do not need to find appropriate strategies to obtain regularization images, and faster, since we do not need additional training iterations for learning from the regularization images.
We also show that the choice of macro class for personalizing a given image affects the performance, e.g., using ``car'' instead of ``Lamborghini'' as the macro class in \Cref{fig:teaser} affects the quality of the outcome (see supplementary).
Based on this, removing the need for regularization images removes an additional dependency and decreases the need for manual selections.

Additionally, many personalization approaches struggle to render specific backgrounds or add new objects often due to some degree of overfitting to the target concept.
For these scenarios, we propose a novel localized attention-guided (LAG) sampling scheme, which allows us to use the finetuned residuals with the original model to generate the target concept and the rest of the image, respectively.
To achieve this, we use the attention maps from the cross-attention layers of the diffusion model at each timestep to predict the location of the concept in the generated image and then apply the features, produced using the personalized residuals, only in the predicted region such that the rest of the image (e.g., background and other objects) is generated by the original model.
Thus, we ensure that we do not lose the capability of generating specific backgrounds or unrelated objects due to overfitting.
Furthermore, this sampling approach does not require any additional training or data, and does not increase sampling time as no additional model evaluations are needed.

We evaluate our approach and sampling technique on the CustomConcept101 dataset \cite{kumari2023multi}, which was specifically designed to evaluate personalization approaches.
We use CLIP and DINO scores to evaluate the text-image alignment (i.e., how well the personalized model can generate the concept in novel scenes and environments) and identity preservation of the personalized model (i.e., how well it can generate the desired concept).
We also perform a user study to evaluate human preference for text-image alignment and identity preservation.
Our results show that our model performs on par or better compared to current state-of-the-art baselines while using significantly fewer parameters, not relying on regularization images, and being faster to train.

To summarize, our key contributions are a novel and more efficient low-rank personalization approach for text-to-image diffusion models that works for arbitrary domains and concepts, uses fewer parameters than previous approaches, does not rely on regularization images and is, therefore, faster and simpler to train.
We also introduce a novel \textit{localized attention-guided (LAG) sampling} approach that allows us to flexibly combine the original pretrained and the finetuned model on the fly to generate different parts of the image, without increasing the sampling time and without requiring additional training or user inputs.
Our user study and quantitative evaluations show that our method performs comparably or better than other baselines, and our proposed sampling approach can address challenges with certain types of recontextualization scenarios, such as background changes.
\section{Related Work}
\label{sec:related_work}

\subsection{Personalization of text-to-image models}
The task of text-to-image personalization was proposed by \cite{gal2022image}, where a few example images of the given concept are used to finetune a ``personalized'' token embedding while all other parameters of the model frozen.
Instead of trying to find an embedding within the existing text conditioning space to represent a concept, DreamBooth \cite{ruiz2023dreambooth} finetunes the diffusion model's parameters to directly inject the concept into the learned prior, leading to better performance.
Custom Diffusion \cite{kumari2023multi} only finetunes the cross-attention weights in addition to the token embedding to achieve more efficient personalization compared to DreamBooth.
Based on these works, other aim to improve the performance and efficiency of personalizing text-to-image models through approaches such as, but not limited to, learning multiple personalized tokens \cite{dong2022dreamartist,hertz2022prompt}, imposing constraints on the trainable parameters (e.g., key-locking \cite{tewel2023key}, orthogonality \cite{qiu2023controlling}, low-rank \cite{smith2023continual}, singular values only \cite{han2023svdiff}), training hypernetworks and domain-specific encoders \cite{ruiz2023hyperdreambooth, xiao2023fastcomposer, li2023blip,gal2023encoder}, and injecting of visual features \cite{wei2023elite,hao2023vico,xiao2023fastcomposer}.

\subsection{Attention-guided text-to-image synthesis} \label{sec:rw-attention}
Attention layers \cite{vaswani2017attention} have been shown to play an important role in the success of text-conditioned image synthesis using diffusion models. Recent works propose to manipulate attention maps from these layers for guided synthesis and editing. \cite{chefer2023attend} modifies cross-attention values to guide the generation process so that the subjects specified in an input prompt appear and the attributes are associated to its corresponding subject. \cite{balaji2022ediffi,he2023localized} enable conditioning on a user-provided layout by guiding the localization of objects via cross-attention manipulation. Given an existing image and a prompt that describes the image, \cite{hertz2022prompt,epstein2023diffusion} synthesize/edit images by manipulating the cross-attention map corresponding to the editing target. Similarly, \cite{brooks2023instructpix2pix} performs edits on existing images albeit through instructions and modifications within self-attention layers.
\section{Approach}

Our method consists of two components: 1) \textit{Personalized residuals}, which encode the identity of a given concept through a set of learned offsets applied to a subset of weights within a pretrained text-to-image diffusion model, and 2) \textit{Localized attention-guided (LAG) sampling}, which leverages attention maps to localize where the residuals are applied, essentially allowing a single image to be efficiently generated by leveraging both the base diffusion model and the personalized residuals.

\subsection{Preliminaries} \label{sec:approach-prelim}

\textbf{Diffusion models.}
Diffusion models \cite{ho2020denoising} consist of a fixed forward noising process that gradually adds noise to an image, and a learned denoising process that iteratively removes noise to produce a valid image. The denoising process is learned through a U-Net \cite{ronneberger2015u} $\epsilon_\theta$, parameterized by $\theta$, and is conditioned on an image $x_t$ noised to timestep $t$, and $t$ itself. Text guidance can be incorporated through conditioning on embeddings $c = \tau(y)$ of input prompts $y$ from a text encoder $\tau$, such as CLIP \cite{radford2021learning}.

In this work, we leverage Stable Diffusion, a text-conditioned latent diffusion model (LDM) \cite{rombach2022high}. An LDM is a variant of a diffusion model that operates in the latent space of a variational autoencoder \cite{kingma2014auto}. The encoder $\mathcal{E}$ embeds an input image $x$ into a latent representation $z=\mathcal{E}(x)$ and a decoder $\mathcal{D}$ maps $z$ back into pixel space $x'=\mathcal{D}(z)$. The diffusion portion of LDM operates on $z$ and is trained using the following objective:

\begin{equation} \label{eq:diffusion}
    \mathcal{L}_{\text{LDM}} = \mathbb{E}_{z\sim\mathcal{E}(x), y, \epsilon\sim\mathcal{N}(0,1), t} \Big[ \| \epsilon - \epsilon_\theta\big(z_t, t, \tau(y)\big) \|_2^2 \Big].
\end{equation}

\textbf{Low rank adaptation (LoRA).}
Low rank adaptation (LoRA) \cite{hu2021lora} is an efficient method originally proposed for updating large language models through learned residuals instead of directly finetuning their parameters. For a given layer of the pretrained model with weight matrix $W_0 \in \mathbb{R}^{m \times n}$, LoRA learns two matrices $A$ and $B$ whose product forms a residual $\Delta W = AB \in \mathbb{R}^{m \times n}$, where $A\in\mathbb{R}^{m \times r}$, $B\in\mathbb{R}^{r \times n}$, and $r\ll\min(m, n)$ is the rank. The updated weight matrix is then defined as $W' = W_0 + \Delta W$. With small values of $r$, LoRA has been shown to significantly reduce the number of learnable parameters while retaining or even improving performance.

\subsection{Learning residuals for capturing identity} \label{sec:approach-residuals}

The goal of personalizing text-to-image models is to faithfully capture the identity of a target concept while simultaneously avoiding overfitting so that the concept can be recontextualized into new settings and configurations. Since concepts are often learned using only a few reference images, directly finetuning the weights of a very large generative model can easily lead to overfitting and/or overwriting unnecessary parts of the learned language prior. Instead we propose to use a LoRA-based approach to learn low-rank offsets for a small subset of the diffusion model weights which will represent the target concept.
Thus, we are able to recover the full generative capacity of the original model by simply not applying the learned residuals at inference.

\begin{figure*}
    \centering
    \includegraphics[width=\textwidth]{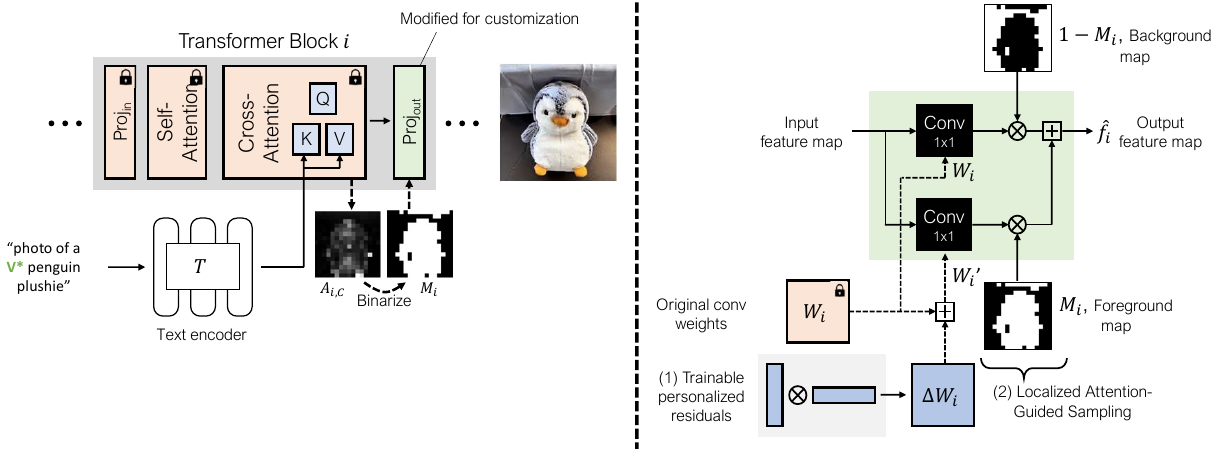}
    \caption{Overview of our proposed work. (1) \textit{Personalized residuals}: We learn low-rank residuals for the output projection layer within each transformer block in the diffusion model. The residuals contain relatively few parameters, are fast to train, and do not require any regularization images during training. (2) \textit{Localized attention-guided sampling}: We optionally apply the personalized residuals only in the areas that the cross-attention layers have localized the concept via predicted attention maps. Thus, we can combine the newly learned concept with the original generative prior of the base diffusion model within a single image.\vspace{-0.5em}}
    \label{fig:overview}
\end{figure*}

The diffusion model contains multiple transformer blocks, which consist of self- and cross-attention layers \cite{vaswani2017attention} with a 1$\times$1 conv projection layer on either end (see \Cref{fig:overview}). While several approaches primarily target the cross-attention layers due to their learning of relationships between text and images, we choose to learn offsets for the output projection conv layers because these localized operations can capture finer details than the global operations of cross-attention.

We illustrate the process of learning personalized residuals in \Cref{fig:overview}. Given a pretrained text-to-image diffusion model containing $L$ transformer blocks, we learn $\Delta W_i = A_i B_i \in \mathbb{R}^{m_i \times m_i}$ for the output projection layer $l_\text{{proj\_out},i}$ with weight matrix $W_i\in\mathbb{R}^{m_i \times m_i \times 1}$ within each transformer block $i$, where $A_i\in\mathbb{R}^{m_i \times r_i}$ and $B_i\in\mathbb{R}^{r_i \times m_i}$. We reshape the residual such that $\Delta W_i\in\mathbb{R}^{m_i \times m_i \times 1}$ and add to the original weights $W_i$ to produce $W_i' = W_i + \Delta W_i$. The $\Delta W_i$'s are updated using the original diffusion objective in \Cref{eq:diffusion}.

Similar to other works, we associate the concept with a unique identifier token (e.g., \texttt{V*}), which is initialized using a rarely occurring token embedding. During training, we use the unique token and macro class of the concept in a fixed template for the prompt associated with each reference image (e.g., ``a photo of a \texttt{V* macro class}'').
%
Personalization approaches that involve direct updates to the diffusion model's weights are susceptible to overwriting parts of the existing generative prior with the new concept and thus explicitly require ``prior preservation'' through regularization images during training \cite{ruiz2023dreambooth,kumari2023multi}. Since our method does not directly update the diffusion model, we avoid this issue entirely and eliminate the burden on the user to determine an effective set of regularization images, which is not always straightforward. Additionally, the low-rank constraint on the residuals reduces the number of trainable parameters, making our method a simpler and more efficient approach for personalization.

\subsection{Localized attention-guided sampling} \label{sec:approach-sampling}

With our residual-based personalization approach, we have additional flexibility in how the offsets are applied at inference. We introduce a new \textit{localized attention-guided} (LAG) sampling method to better combine a newly learned concept with the original generative prior of the diffusion model. As shown in \Cref{fig:overview}, within every transformer block of the diffusion model is a cross-attention layer, which aims to learn the correspondence between text tokens and image regions. Each cross-attention layer computes attention maps $A_{y_i}$ for each token $y_i$ in the prompt, indicating where the token will affect the generated image. The attention maps are produced using the following equation:

\begin{equation}
    A(Q, K) = \text{softmax}\Big(\frac{QK^\top}{\sqrt{d_k}}\Big),
\end{equation}

where $Q=W^Q x$ is the query, $K=W^K y$ is the key, and $d_k$ is the dimension of the query and key.

Given the indices $\mathcal{C}$ of the unique identifier and macro class tokens specifying the concept (e.g., ``\texttt{V*}'' and ``dog''), we sum the values of the corresponding attention maps $A_{i, \mathcal{C}} = \sum_{j\in\mathcal{C}} A_j$ in transformer block $i$, and then binarize using its median value to get $M_i = \text{binarize}(A_{i, \mathcal{C}})$. Finally, we compute the output feature $\hat{f}_i$ of each transformer block $i$ as:

\begin{equation}
    \hat{f}_i = (1 - M_i) \otimes f_i + M_i \otimes f_i',
\end{equation}

where $f_i = W_i x$ is the feature produced using the original conv weight $W_i$, and $f_i' = W_i' x$ is the feature produced using the updated weight from the personalized residual $W_i' = W_i + \Delta W_i$. Thus, the identity represented through the personalized residuals is only being applied in the regions corresponding to the target concept, and the remaining regions are generated by the original diffusion model. The proposed LAG sampling technique is visualized in \Cref{fig:sampling-comparison}. 

While there exist personalization works using attention guidance (e.g., \cite{xiao2023fastcomposer,hao2023vico}), they often rely on object masks and/or additional losses at train time to focus on the relevant object location in the reference images, whereas manually-provided object masks or specific training are not needed to enable LAG. Additionally, LAG sampling explicitly merges the features of two layers (personalized/finetuned and original/non-finetuned) on-the-fly based on the cross-attention maps obtained during inference and has negligible impact on the sampling speed. In contrast, other synthesis/editing works (see \Cref{sec:rw-attention}) use cross-attention values to up- or down-weight the influence of specific tokens at specific image locations.

LAG sampling can be beneficial in scenarios where the learned residuals overfit to the reference images and have not effectively disentangled the target concept from the background, which can occur as a consequence of ambiguities of the target concept given the reference images or model biases (e.g., furniture often photographed indoors). By leveraging the attention maps from the tokens denoting the concept, we can localize the residuals so that they do not affect the background, which can instead be generated using the base model.
\section{Experiments} \label{sec:experiments}

In this section, we describe our experimental setup and evaluation protocols, and visualize examples using the proposed personalized residuals with and without localized attention-guided sampling.

\subsection{Training details} \label{sec:exp-training}

We build upon Stable Diffusion v1.4 \cite{rombach2022high}. For each transformer block $i$, we compute the rank $r_i$ for its output projection convolution layer with weight matrix $W_i\in\mathbb{R}^{m_i \times m_i \times 1}$ as $r_i = 0.05m_i$, totalling 1.2M trainable parameters ($\sim$0.1\% of Stable Diffusion). Each of the low-rank matrices are randomly initialized. We train our method for 150 iterations with a batch size of 4 and learning rate of 1.0e-3 on 1 A100 GPU ($\sim$3 minutes) across all experiments.

\subsection{Baselines} \label{sec:exp-baselines}

We focus on comparisons to open-domain (i.e., does not require encoders limited to a single given domain) approaches with publicly available code. Specifically, we compare our method against four baselines: Textual Inversion \cite{gal2022image}, DreamBooth \cite{ruiz2023dreambooth}, Custom Diffusion \cite{kumari2023multi}, and ViCo \cite{hao2023vico}. Textual Inversion freezes the entire diffusion model and optimizes only the unique identifier token \texttt{V*} for each concept. ViCo optimizes \texttt{V*} as well as newly added cross-attention layers to the diffusion model to incorporate visual information from the reference images while keeping the rest of the model frozen. DreamBooth finetunes the entire diffusion model using the reference images and a set of regularization images, which are generated within the same domain as the target concept using the original model. While DreamBooth was originally proposed using Imagen \cite{saharia2022photorealistic}, we use an open-source version built on Stable Diffusion\footnote{\url{https://github.com/XavierXiao/Dreambooth-Stable-Diffusion}}. Custom Diffusion finetunes only the key and value weights of the cross-attention layers in addition to the identifier token embedding, and uses a set of real regularization images sampled from LAION-400M \cite{schuhmann2021laion}.

We use the recommended settings described by each paper. For Textual Inversion and ViCo, which initialize the identifier token embedding to a single word that best represents the concept, we use our best discretion to pick a word most similar to the macro class given by CustomConcept101.

\subsection{Evaluation metrics} \label{sec:exp-metrics}

Following the protocol described in \cite{kumari2023multi}, we leverage the CustomConcept101 dataset, consisting of 101 concepts across 16 broader categories. For every concept we generate 50 samples for each of the 20 prompts given by the dataset. We use DDIM sampling \cite{song2020denoising} with $N=50$ steps, $\eta=0.0$, and a guidance scale of 6.0 for all methods. We set the same random seed for sampling across each method so that the ``choice'' of starting noise does not impact the results. Results of our method with LAG sampling are explicitly labeled as such.

We evaluate each method for text alignment and image alignment. \textit{Text alignment} is measured as the similarity between the CLIP \cite{radford2021learning} text feature of the input prompt and the CLIP image feature of the resulting generated image. \textit{Image alignment} is measured as the similarity between image features from either CLIP or DINO \cite{caron2021emerging} of the reference images and corresponding generated images.

Additionally, we evaluate both text and image alignment using human evaluations through user studies on Amazon Mechanical Turk (AMT). For each text alignment case, we display a text prompt and a pair of corresponding generated images, and ask users \textit{``Which image is more consistent with the given text prompt?''}. For each image alignment case, we display 3 reference images for a concept and a pair of corresponding generated images, and ask \textit{``Which image better preserves the identity of the subject in the provided reference images?''}. For both studies, each pair of images contains one from \{Textual Inversion, ViCo, DreamBooth, Custom Diffusion, Ours w/ LAG sampling\} and one from ours with normal DDIM sampling. Users can select either image or neither (\textit{``Not sure''}).

\subsection{Results} \label{sec:exp-results}

\begin{figure*}
    \centering
    \includegraphics[width=\textwidth]{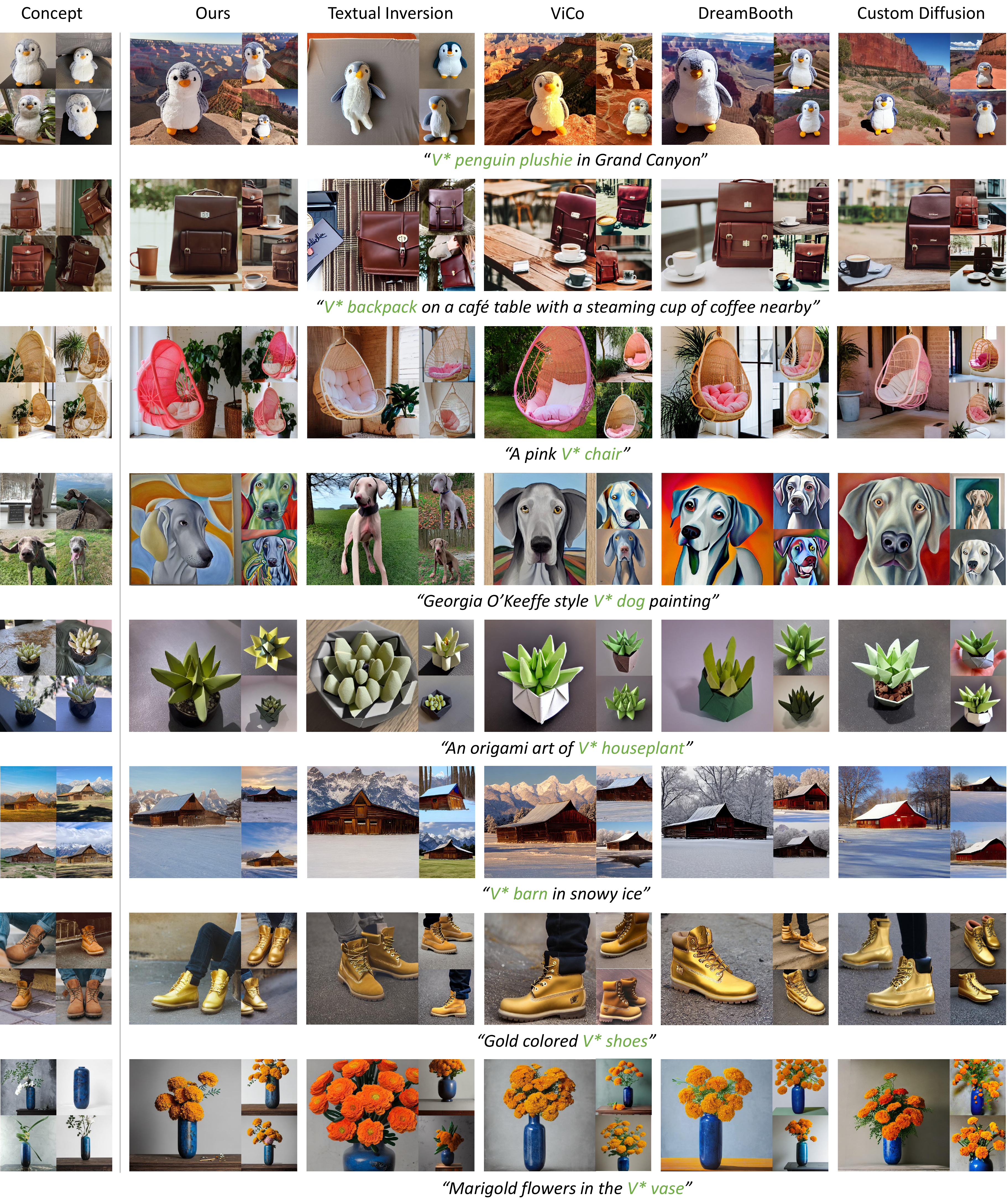}
    \caption{Qualitative comparison of our proposed approach with the baselines.}
    \label{fig:qualitative-comparison}
\end{figure*}

\begin{table}[t]
\caption{Quantitative evaluations for text and image alignment using the similarity of CLIP and DINO features. We report the number of parameters for each method in addition to scores from the base Stable Diffusion model, which is not trained for personalization, for reference.}
\label{tab:quantitative-evals}
\begin{adjustbox}{width=\linewidth}
\begin{tabular}{lcccc}
\multicolumn{1}{l}{\bf Method}  &  \multicolumn{1}{c}{\bf \# params}  &  \multicolumn{1}{c}{\bf CLIP text}  &  \multicolumn{1}{c}{\bf CLIP image}  &  \multicolumn{1}{c}{\bf DINO image} \\
\hline
Textual Inversion  &  768  &  0.6150  &  0.7259  &  0.4700 \\
ViCo  &  51.3M  &  0.7403  &  0.7111  &  0.4678 \\
DreamBooth  &  983M  &  0.7536  &  0.7424  &  0.5212  \\
Custom Diffusion  &  19M  &  \textbf{0.7664}  &  0.7074  &  0.4669  \\
Ours  & 1.2M  &  0.7193 &  \textbf{0.7594}  &  \textbf{0.5671}  \\
Ours w/ LAG sampling  & 1.2M &  0.7220  &  0.7424  &  0.5411 \\
\hdashline
Stable Diffusion  & 983M &  0.8126  &  0.6207  &  0.2920  \\
\end{tabular}\vspace{-0.5em}
\end{adjustbox}
\end{table}

\begin{table}[t]
\caption{Human preference evaluations for text and image alignment through Amazon Mechanical Turk. We perform bootstrap resampling over the 1250 responses collected for each task.}
\label{tab:amt}
\begin{adjustbox}{width=\linewidth}
\begin{tabular}{lccccc}
\multirow{2}{*}{\bf Ours vs.}  &  \textbf{Textual}  &  \multirow{2}{*}{\bf ViCo}  &  \multirow{2}{*}{\bf DreamBooth}  &  \textbf{Custom}  &  \textbf{Ours w/} \\
  &  \textbf{Inversion}  &  &  &  \textbf{Diffusion}  &  \textbf{LAG} \\
\hline
\multirow{2}{*}{Text}  &  \textbf{81.85}  &  37.40  &  41.34  &  \textbf{50.99}  &  \textbf{58.57} \\
  &  $\pm$4.15\%  &  $\pm$7.46\%  &  $\pm$5.08\%  &  $\pm$5.46\%  &  $\pm$6.34\%  \\
\midrule
\multirow{2}{*}{Image}  &  \textbf{61.96}  &  \textbf{62.11}  &  \textbf{51.33}  &  \textbf{63.27}  &  26.26 \\
  &  $\pm$4.76\%  &  $\pm$5.80\%  &  $\pm$4.65\%  &  $\pm$5.59\%  &  $\pm$4.91\%
\end{tabular}
\end{adjustbox}
\end{table}

\begin{figure*}
    \centering
    \begin{subfigure}[T]{0.49\textwidth}
        \centering
        \includegraphics[width=\textwidth]{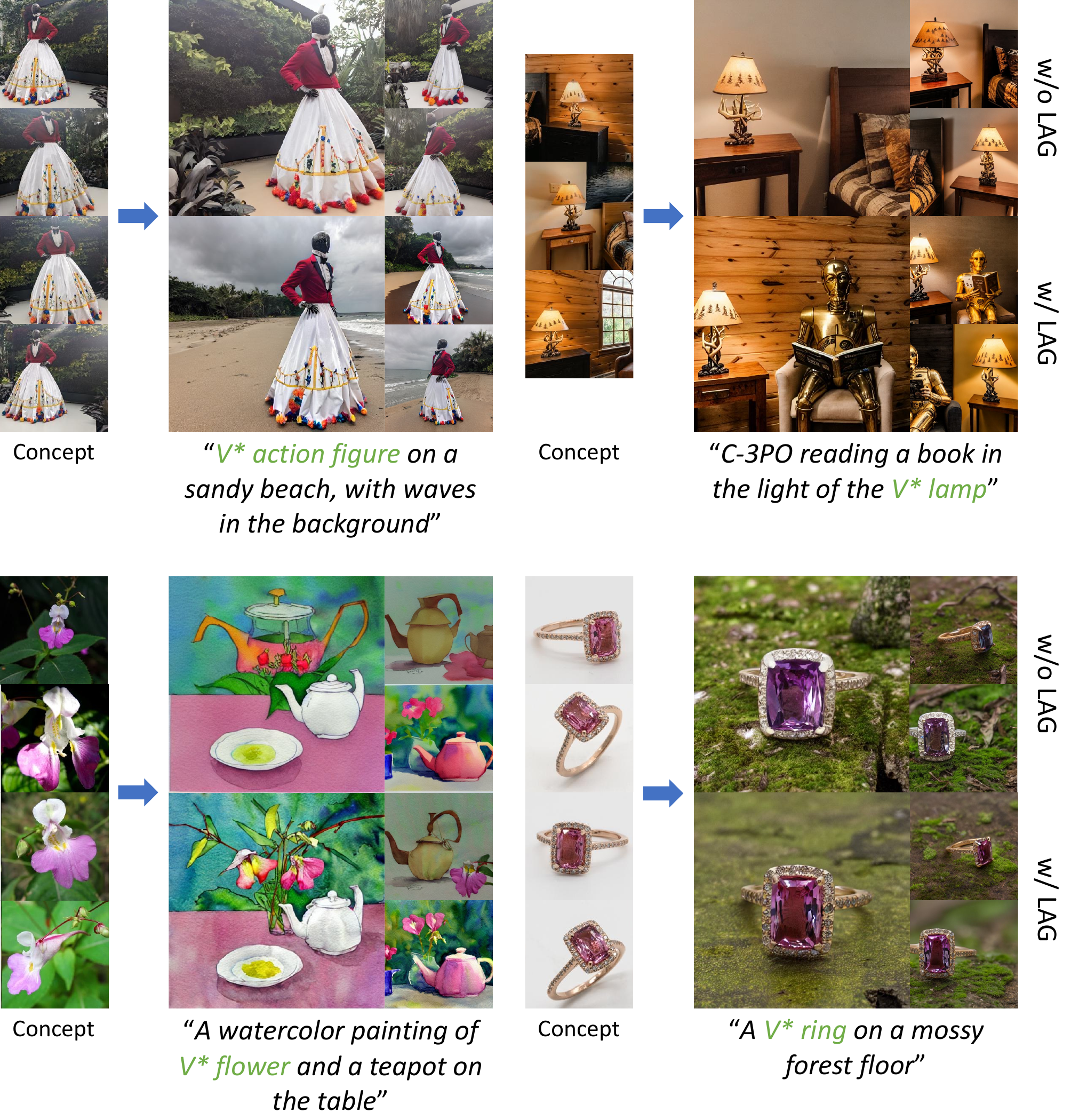}
        \caption{Examples where LAG produces results that are better aligned with the concept and prompt.}
        \label{fig:lag-vs-normal}
    \end{subfigure}
    \hfill
    \begin{subfigure}[T]{0.49\textwidth}
        \center
        \includegraphics[width=\textwidth]{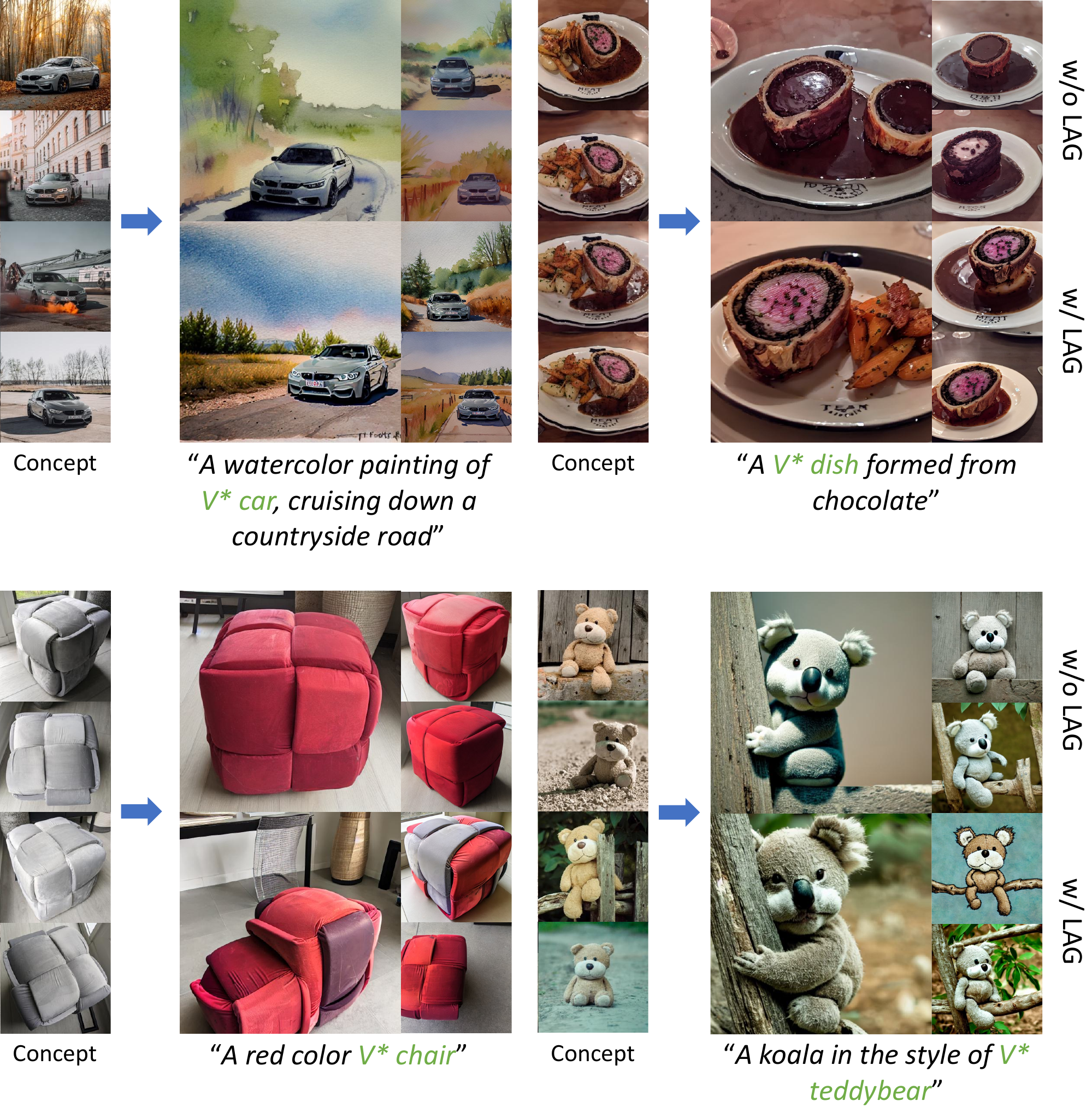}
        \vspace{-1.6mm}\caption{Examples where normal sampling produces results that are better aligned with the concept and prompt.}
        \label{fig:normal-vs-lag}
    \end{subfigure}
    \caption{Comparison of image generated with and without LAG sampling. We use the same starting noise map to generate corresponding pairs of images to directly visualize how LAG sampling affects the output image.\vspace{-0.5em}}
    \label{fig:sampling-comparison}
\end{figure*}

We visualize samples generated by each method for various types of prompts in \Cref{fig:qualitative-comparison}. Textual Inversion fails to reliably capture the concept's identity and/or the prompt whereas all other methods, including ours, are able to better preserve the concept's identity while also adhering to the prompt. We highlight that our method is able to achieve these results while having significantly fewer learnable parameters and requiring less training time compared to ViCo, DreamBooth, and Custom Diffusion, as well as not leveraging regularization images.

We compare examples using our proposed personalized residuals with and without localized attention-guided sampling in \Cref{fig:sampling-comparison}. We illustrate how LAG sampling affects the output image by using the same starting noise map $z_T$ to sample each pair of \{w/o LAG, w/ LAG\} images. 
We highlight scenarios where LAG sampling performs better than normal sampling in \Cref{fig:lag-vs-normal} and vice versa in \Cref{fig:normal-vs-lag}. 

Quantitative evaluations for text and image alignment using CLIP and DINO are shown in \Cref{tab:quantitative-evals}. We include results using the original Stable Diffusion model, which has no notion of any of the concepts, for reference. We show that our method performs similarly with and without LAG sampling averaged across the whole dataset, demonstrating higher image alignment and slightly lower text alignment than the more computationally-heavy baselines.

However, as seen by the results of 1250 responses collected through AMT user studies for both text and image alignment in \Cref{tab:amt}, we show that the CLIP text alignment scores do not necessarily correlate to human preference. We observe that our method performs similarly to Custom Diffusion for text alignment, which was assigned the highest CLIP text score, and outperforms all baselines for image alignment. Again, we note that our method achieves similar performance to the better performing baselines while being significantly more computationally efficient. We also compare our method with and without LAG sampling in the user studies and show that LAG is preferred for image alignment but not text alignment. Further analysis comparing the two sampling approaches can be found in the supplementary.

We also train and evaluate our method using CLIP similarity to select the ``most representative'' macro class among the 117k nouns in WordNet \cite{miller1995wordnet} for each concept. In \Cref{tab:macro-comparison}, we show that using the WordNet macro class leads to further improvements in image alignment while decreasing text alignment, the latter of which may not necessarily reflect human preference as previously demonstrated. See the supplementary for additional discussions.


\textbf{Ablation studies.} We perform ablation studies on changing the targets for where the residuals are applied, removing the macro class from the prompt, including regularization images (sampled from LAION) during training, updating the concept identifier token embedding \texttt{V*}, and varying the rank of the residuals. Results are shown in \Cref{tab:ablation-others} (see \Cref{tab:ablation-rank} for results on changing the rank).

\begin{table}[t]
\caption{We evaluate our method using two different targets for the residuals and altering various training settings.}
\label{tab:ablation-others}
\begin{adjustbox}{width=\linewidth}
\begin{tabular}{lccc}
\multicolumn{1}{l}{\bf Method}  &  \multicolumn{1}{c}{\bf CLIP text}  &  \multicolumn{1}{c}{\bf CLIP image}  &  \multicolumn{1}{c}{\bf DINO image} \\
\hline
KV weights  &  0.7172  &  0.7508  &  0.5353 \\
$l_\text{proj\_in}$ weights  &  0.7136  &  0.7460  &  0.5333 \\
KV + $l_\text{proj\_out}$ weights  &  0.7049  &  0.7733  &  0.5868 \\
KV + $l_\text{proj\_in}$ + $l_\text{proj\_out}$ weights  &  0.6739  &  0.7870  &  0.6040 \\
w/o macro class  &  0.6605  &  0.6521  &  0.3798 \\
w/ reg images  &  0.7204  &  0.6771  &  0.3830 \\
Update token embedding  &  0.6673  &  0.8000  &  0.6194 \\
\hdashline
Ours  &  0.7193  &  0.7594  &  0.5671
\end{tabular}\vspace{-0.5em}
\end{adjustbox}
\end{table}

We show that changing where the residuals are applied to either the key and value weights of the cross-attention layers (like Custom Diffusion) or the input projection conv layer (rather than the output) slightly decreases the scores across all three metrics compared to our proposed approach. We hypothesize that the output projection layer achieves noticeably higher identity preservation because it refines the feature map at the end of each block. Additionally, learning residuals for multiple layers simultaneously leads to overfitting to the reference images as demonstrated by the higher image alignment scores and lower text alignment.

Omitting the macro class leads to significant drops across all metrics, demonstrating that the additional information is useful to our method for knowing what within the reference images is important to model. Similar to the effect of using regularization images for DreamBooth and Custom Diffusion, regularization images slightly improves text alignment but decreases image alignment. On the other hand, updating the token embedding for \texttt{V*} leads to overfitting as shown by the increase in image alignment and decrease in text alignment.

\section{Conclusion} \label{sec:conclusion}

We introduce personalized residuals, a method for concept-driven synthesis using text-to-image diffusion models. Previous approaches to personalization are often slow to train, have high computational demands, require regularization images, and/or have difficulty recontextualizing the target concept. Through our proposed LoRA-based approach that learns a small set of residuals to represent the identity of a concept, we reduce the number of learnable parameters and training time and remove the reliance on domain regularization while maintaining flexibility with editing. We also introduce localized attention-guided sampling which applies the personalized residuals only in regions where the concept is localized via the cross-attention mechanism. We evaluate our method across several metrics to show that we are able to efficiently enable personalization.

\textbf{Limitations and future work.} We show that localized sampling is not always the best choice (e.g., changing the color of a concept) and relies on the cross-attention layers to produce high-quality attention maps, which is not always the case. Our approach can be sensitive to the choice of macro class and inherits the pretrained model's biases and limitations, such as mixing up the relationship between attributes in the prompt. Finally, we leave multi-concept generation through LAG sampling as future work.


{
    \small
    \bibliographystyle{ieeenat_fullname}
    \bibliography{main}

\begin{thebibliography}{33}
\providecommand{\natexlab}[1]{#1}
\providecommand{\url}[1]{\texttt{#1}}
\expandafter\ifx\csname urlstyle\endcsname\relax
  \providecommand{\doi}[1]{doi: #1}\else
  \providecommand{\doi}{doi: \begingroup \urlstyle{rm}\Url}\fi

\bibitem[Balaji et~al.(2022)Balaji, Nah, Huang, Vahdat, Song, Kreis, Aittala, Aila, Laine, Catanzaro, et~al.]{balaji2022ediffi}
Yogesh Balaji, Seungjun Nah, Xun Huang, Arash Vahdat, Jiaming Song, Karsten Kreis, Miika Aittala, Timo Aila, Samuli Laine, Bryan Catanzaro, et~al.
\newblock ediffi: Text-to-image diffusion models with an ensemble of expert denoisers.
\newblock \emph{arXiv preprint arXiv:2211.01324}, 2022.

\bibitem[Brooks et~al.(2023)Brooks, Holynski, and Efros]{brooks2023instructpix2pix}
Tim Brooks, Aleksander Holynski, and Alexei~A Efros.
\newblock Instructpix2pix: Learning to follow image editing instructions.
\newblock In \emph{Proceedings of the IEEE/CVF Conference on Computer Vision and Pattern Recognition}, pages 18392--18402, 2023.

\bibitem[Caron et~al.(2021)Caron, Touvron, Misra, J{\'e}gou, Mairal, Bojanowski, and Joulin]{caron2021emerging}
Mathilde Caron, Hugo Touvron, Ishan Misra, Herv{\'e} J{\'e}gou, Julien Mairal, Piotr Bojanowski, and Armand Joulin.
\newblock Emerging properties in self-supervised vision transformers.
\newblock In \emph{Proceedings of the IEEE/CVF international conference on computer vision}, pages 9650--9660, 2021.

\bibitem[Chefer et~al.(2023)Chefer, Alaluf, Vinker, Wolf, and Cohen-Or]{chefer2023attend}
Hila Chefer, Yuval Alaluf, Yael Vinker, Lior Wolf, and Daniel Cohen-Or.
\newblock Attend-and-excite: Attention-based semantic guidance for text-to-image diffusion models.
\newblock \emph{ACM Transactions on Graphics (TOG)}, 42\penalty0 (4):\penalty0 1--10, 2023.

\bibitem[Dong et~al.(2022)Dong, Wei, and Lin]{dong2022dreamartist}
Ziyi Dong, Pengxu Wei, and Liang Lin.
\newblock Dreamartist: Towards controllable one-shot text-to-image generation via contrastive prompt-tuning.
\newblock \emph{arXiv preprint arXiv:2211.11337}, 2022.

\bibitem[Epstein et~al.(2023)Epstein, Jabri, Poole, Efros, and Holynski]{epstein2023diffusion}
Dave Epstein, Allan Jabri, Ben Poole, Alexei~A Efros, and Aleksander Holynski.
\newblock Diffusion self-guidance for controllable image generation.
\newblock \emph{arXiv preprint arXiv:2306.00986}, 2023.

\bibitem[Gal et~al.(2022)Gal, Alaluf, Atzmon, Patashnik, Bermano, Chechik, and Cohen-Or]{gal2022image}
Rinon Gal, Yuval Alaluf, Yuval Atzmon, Or Patashnik, Amit~H Bermano, Gal Chechik, and Daniel Cohen-Or.
\newblock An image is worth one word: Personalizing text-to-image generation using textual inversion.
\newblock \emph{arXiv preprint arXiv:2208.01618}, 2022.

\bibitem[Gal et~al.(2023)Gal, Arar, Atzmon, Bermano, Chechik, and Cohen-Or]{gal2023encoder}
Rinon Gal, Moab Arar, Yuval Atzmon, Amit~H Bermano, Gal Chechik, and Daniel Cohen-Or.
\newblock Encoder-based domain tuning for fast personalization of text-to-image models.
\newblock \emph{ACM Transactions on Graphics (TOG)}, 42\penalty0 (4):\penalty0 1--13, 2023.

\bibitem[Han et~al.(2023)Han, Li, Zhang, Milanfar, Metaxas, and Yang]{han2023svdiff}
Ligong Han, Yinxiao Li, Han Zhang, Peyman Milanfar, Dimitris Metaxas, and Feng Yang.
\newblock Svdiff: Compact parameter space for diffusion fine-tuning.
\newblock \emph{arXiv preprint arXiv:2303.11305}, 2023.

\bibitem[Hao et~al.(2023)Hao, Han, Zhao, and Wong]{hao2023vico}
Shaozhe Hao, Kai Han, Shihao Zhao, and Kwan-Yee~K Wong.
\newblock Vico: Detail-preserving visual condition for personalized text-to-image generation.
\newblock \emph{arXiv preprint arXiv:2306.00971}, 2023.

\bibitem[He et~al.(2023)He, Salakhutdinov, and Kolter]{he2023localized}
Yutong He, Ruslan Salakhutdinov, and J~Zico Kolter.
\newblock Localized text-to-image generation for free via cross attention control.
\newblock \emph{arXiv preprint arXiv:2306.14636}, 2023.

\bibitem[Hertz et~al.(2022)Hertz, Mokady, Tenenbaum, Aberman, Pritch, and Cohen-Or]{hertz2022prompt}
Amir Hertz, Ron Mokady, Jay Tenenbaum, Kfir Aberman, Yael Pritch, and Daniel Cohen-Or.
\newblock Prompt-to-prompt image editing with cross attention control.
\newblock \emph{arXiv preprint arXiv:2208.01626}, 2022.

\bibitem[Ho et~al.(2020)Ho, Jain, and Abbeel]{ho2020denoising}
Jonathan Ho, Ajay Jain, and Pieter Abbeel.
\newblock Denoising diffusion probabilistic models.
\newblock \emph{Advances in neural information processing systems}, 33:\penalty0 6840--6851, 2020.

\bibitem[Hu et~al.(2021)Hu, Shen, Wallis, Allen-Zhu, Li, Wang, Wang, and Chen]{hu2021lora}
Edward~J Hu, Yelong Shen, Phillip Wallis, Zeyuan Allen-Zhu, Yuanzhi Li, Shean Wang, Lu Wang, and Weizhu Chen.
\newblock Lora: Low-rank adaptation of large language models.
\newblock \emph{arXiv preprint arXiv:2106.09685}, 2021.

\bibitem[Kingma and Welling(2014)]{kingma2014auto}
Diederik~P Kingma and Max Welling.
\newblock Auto-encoding variational bayes.
\newblock \emph{International Conference on Learning Representations (ICLR)}, 2014.

\bibitem[Kumari et~al.(2023)Kumari, Zhang, Zhang, Shechtman, and Zhu]{kumari2023multi}
Nupur Kumari, Bingliang Zhang, Richard Zhang, Eli Shechtman, and Jun-Yan Zhu.
\newblock Multi-concept customization of text-to-image diffusion.
\newblock In \emph{Proceedings of the IEEE/CVF Conference on Computer Vision and Pattern Recognition}, pages 1931--1941, 2023.

\bibitem[Li et~al.(2023)Li, Li, and Hoi]{li2023blip}
Dongxu Li, Junnan Li, and Steven~CH Hoi.
\newblock Blip-diffusion: Pre-trained subject representation for controllable text-to-image generation and editing.
\newblock \emph{arXiv preprint arXiv:2305.14720}, 2023.

\bibitem[Miller(1995)]{miller1995wordnet}
George~A Miller.
\newblock Wordnet: a lexical database for english.
\newblock \emph{Communications of the ACM}, 38\penalty0 (11):\penalty0 39--41, 1995.

\bibitem[Qiu et~al.(2023)Qiu, Liu, Feng, Xue, Feng, Liu, Zhang, Weller, and Sch{\"o}lkopf]{qiu2023controlling}
Zeju Qiu, Weiyang Liu, Haiwen Feng, Yuxuan Xue, Yao Feng, Zhen Liu, Dan Zhang, Adrian Weller, and Bernhard Sch{\"o}lkopf.
\newblock Controlling text-to-image diffusion by orthogonal finetuning.
\newblock \emph{arXiv preprint arXiv:2306.07280}, 2023.

\bibitem[Radford et~al.(2021)Radford, Kim, Hallacy, Ramesh, Goh, Agarwal, Sastry, Askell, Mishkin, Clark, et~al.]{radford2021learning}
Alec Radford, Jong~Wook Kim, Chris Hallacy, Aditya Ramesh, Gabriel Goh, Sandhini Agarwal, Girish Sastry, Amanda Askell, Pamela Mishkin, Jack Clark, et~al.
\newblock Learning transferable visual models from natural language supervision.
\newblock In \emph{International conference on machine learning}, pages 8748--8763. PMLR, 2021.

\bibitem[Ramesh et~al.(2022)Ramesh, Dhariwal, Nichol, Chu, and Chen]{ramesh2022hierarchical}
Aditya Ramesh, Prafulla Dhariwal, Alex Nichol, Casey Chu, and Mark Chen.
\newblock Hierarchical text-conditional image generation with clip latents.
\newblock \emph{arXiv preprint arXiv:2204.06125}, 1\penalty0 (2):\penalty0 3, 2022.

\bibitem[Rombach et~al.(2022)Rombach, Blattmann, Lorenz, Esser, and Ommer]{rombach2022high}
Robin Rombach, Andreas Blattmann, Dominik Lorenz, Patrick Esser, and Bj{\"o}rn Ommer.
\newblock High-resolution image synthesis with latent diffusion models.
\newblock In \emph{Proceedings of the IEEE/CVF conference on computer vision and pattern recognition}, pages 10684--10695, 2022.

\bibitem[Ronneberger et~al.(2015)Ronneberger, Fischer, and Brox]{ronneberger2015u}
Olaf Ronneberger, Philipp Fischer, and Thomas Brox.
\newblock U-net: Convolutional networks for biomedical image segmentation.
\newblock In \emph{Medical Image Computing and Computer-Assisted Intervention--MICCAI 2015: 18th International Conference, Munich, Germany, October 5-9, 2015, Proceedings, Part III 18}, pages 234--241. Springer, 2015.

\bibitem[Ruiz et~al.(2023{\natexlab{a}})Ruiz, Li, Jampani, Pritch, Rubinstein, and Aberman]{ruiz2023dreambooth}
Nataniel Ruiz, Yuanzhen Li, Varun Jampani, Yael Pritch, Michael Rubinstein, and Kfir Aberman.
\newblock Dreambooth: Fine tuning text-to-image diffusion models for subject-driven generation.
\newblock In \emph{Proceedings of the IEEE/CVF Conference on Computer Vision and Pattern Recognition}, pages 22500--22510, 2023{\natexlab{a}}.

\bibitem[Ruiz et~al.(2023{\natexlab{b}})Ruiz, Li, Jampani, Wei, Hou, Pritch, Wadhwa, Rubinstein, and Aberman]{ruiz2023hyperdreambooth}
Nataniel Ruiz, Yuanzhen Li, Varun Jampani, Wei Wei, Tingbo Hou, Yael Pritch, Neal Wadhwa, Michael Rubinstein, and Kfir Aberman.
\newblock Hyperdreambooth: Hypernetworks for fast personalization of text-to-image models.
\newblock \emph{arXiv preprint arXiv:2307.06949}, 2023{\natexlab{b}}.

\bibitem[Saharia et~al.(2022)Saharia, Chan, Saxena, Li, Whang, Denton, Ghasemipour, Gontijo~Lopes, Karagol~Ayan, Salimans, et~al.]{saharia2022photorealistic}
Chitwan Saharia, William Chan, Saurabh Saxena, Lala Li, Jay Whang, Emily~L Denton, Kamyar Ghasemipour, Raphael Gontijo~Lopes, Burcu Karagol~Ayan, Tim Salimans, et~al.
\newblock Photorealistic text-to-image diffusion models with deep language understanding.
\newblock \emph{Advances in Neural Information Processing Systems}, 35:\penalty0 36479--36494, 2022.

\bibitem[Schuhmann et~al.(2021)Schuhmann, Vencu, Beaumont, Kaczmarczyk, Mullis, Katta, Coombes, Jitsev, and Komatsuzaki]{schuhmann2021laion}
Christoph Schuhmann, Richard Vencu, Romain Beaumont, Robert Kaczmarczyk, Clayton Mullis, Aarush Katta, Theo Coombes, Jenia Jitsev, and Aran Komatsuzaki.
\newblock Laion-400m: Open dataset of clip-filtered 400 million image-text pairs.
\newblock \emph{arXiv preprint arXiv:2111.02114}, 2021.

\bibitem[Smith et~al.(2023)Smith, Hsu, Zhang, Hua, Kira, Shen, and Jin]{smith2023continual}
James~Seale Smith, Yen-Chang Hsu, Lingyu Zhang, Ting Hua, Zsolt Kira, Yilin Shen, and Hongxia Jin.
\newblock Continual diffusion: Continual customization of text-to-image diffusion with c-lora.
\newblock \emph{arXiv preprint arXiv:2304.06027}, 2023.

\bibitem[Song et~al.(2020)Song, Meng, and Ermon]{song2020denoising}
Jiaming Song, Chenlin Meng, and Stefano Ermon.
\newblock Denoising diffusion implicit models.
\newblock \emph{arXiv preprint arXiv:2010.02502}, 2020.

\bibitem[Tewel et~al.(2023)Tewel, Gal, Chechik, and Atzmon]{tewel2023key}
Yoad Tewel, Rinon Gal, Gal Chechik, and Yuval Atzmon.
\newblock Key-locked rank one editing for text-to-image personalization.
\newblock In \emph{ACM SIGGRAPH 2023 Conference Proceedings}, pages 1--11, 2023.

\bibitem[Vaswani et~al.(2017)Vaswani, Shazeer, Parmar, Uszkoreit, Jones, Gomez, Kaiser, and Polosukhin]{vaswani2017attention}
Ashish Vaswani, Noam Shazeer, Niki Parmar, Jakob Uszkoreit, Llion Jones, Aidan~N Gomez, {\L}ukasz Kaiser, and Illia Polosukhin.
\newblock Attention is all you need.
\newblock \emph{Advances in neural information processing systems}, 30, 2017.

\bibitem[Wei et~al.(2023)Wei, Zhang, Ji, Bai, Zhang, and Zuo]{wei2023elite}
Yuxiang Wei, Yabo Zhang, Zhilong Ji, Jinfeng Bai, Lei Zhang, and Wangmeng Zuo.
\newblock Elite: Encoding visual concepts into textual embeddings for customized text-to-image generation.
\newblock \emph{arXiv preprint arXiv:2302.13848}, 2023.

\bibitem[Xiao et~al.(2023)Xiao, Yin, Freeman, Durand, and Han]{xiao2023fastcomposer}
Guangxuan Xiao, Tianwei Yin, William~T Freeman, Fr{\'e}do Durand, and Song Han.
\newblock Fastcomposer: Tuning-free multi-subject image generation with localized attention.
\newblock \emph{arXiv preprint arXiv:2305.10431}, 2023.

\end{thebibliography}
}

\clearpage
\setcounter{page}{1}
\maketitlesupplementary

\section{Additional experimental results}

\begin{figure}
    \centering
    \includegraphics[width=0.9\linewidth]{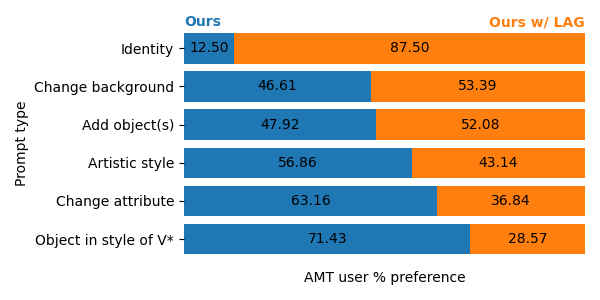}
    \caption{AMT text alignment scores per prompt type.}
    \label{fig:clip-by-prompt}\vspace{-1em}
\end{figure}

We explore the difference in normal and LAG sampling by using ChatGPT to categorize each prompt into \{\textit{add object(s)}, \textit{artistic style}, \textit{change attribute}, \textit{change background}, \textit{identity}, \textit{object in style of \texttt{V*}}\}. We note that a prompt may fall into multiple categories, but we only use one as determined by ChatGPT. We split the AMT evaluations for text alignment by category in \Cref{fig:clip-by-prompt}. We observe that LAG sampling performs best for \textit{identity}, \textit{change background}, and \textit{add object(s)}, which are tasks in which the target object is somewhat independent of the rest of the image. Tasks that require modifying the target (\textit{artistic style}, \textit{change attribute}, \textit{object in style of \texttt{V*}}) perform better with normal DDIM sampling.

\begin{figure*}
    \centering
    \includegraphics[height=0.42\textheight]{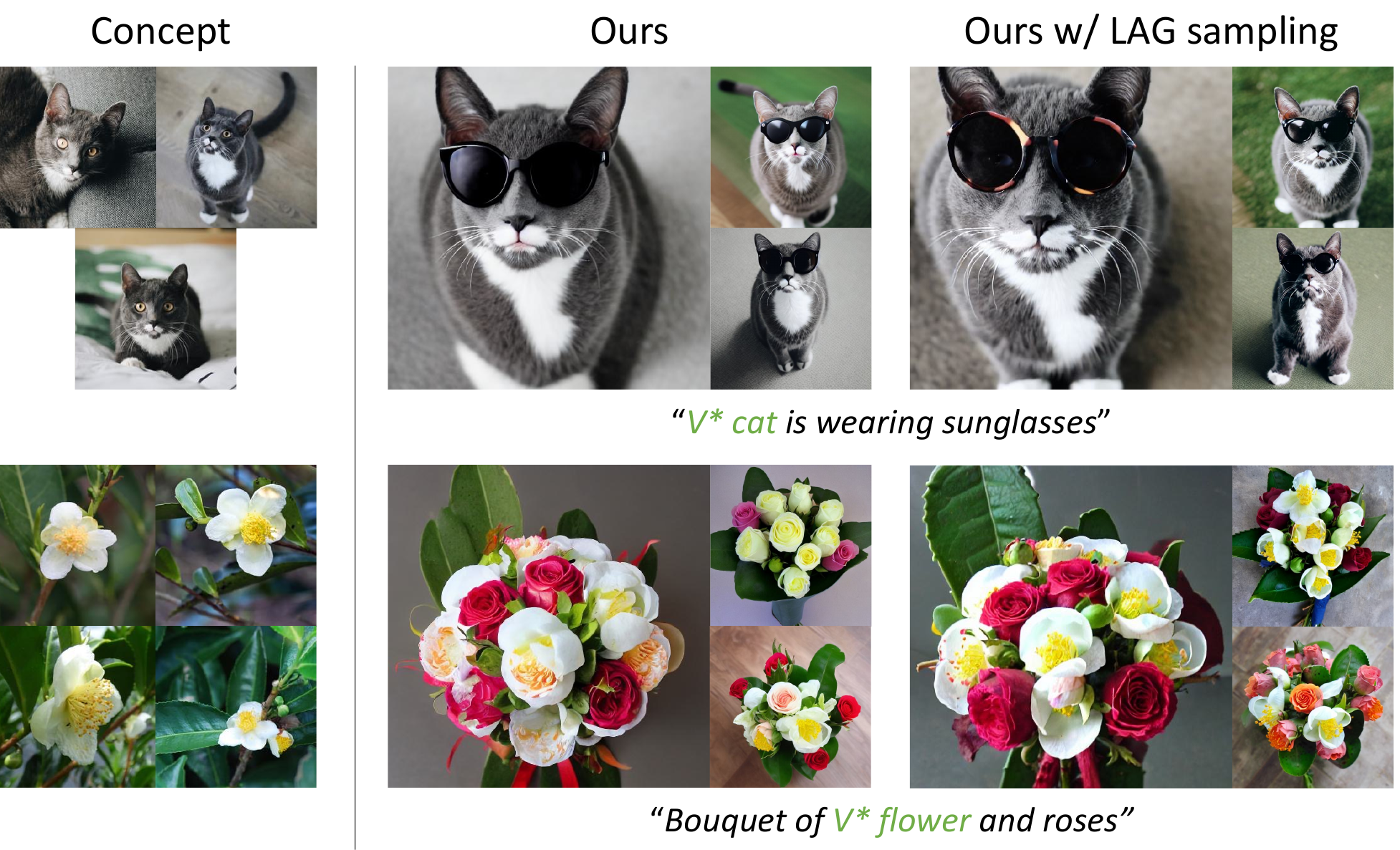}
    \caption{Samples for \textit{add object(s)} prompts using personalized residuals with and without LAG sampling where corresponding pairs are generated using the same input noise map.}
    \label{fig:add-object}
\end{figure*}

\begin{figure*}
    \centering
    \includegraphics[height=0.42\textheight]{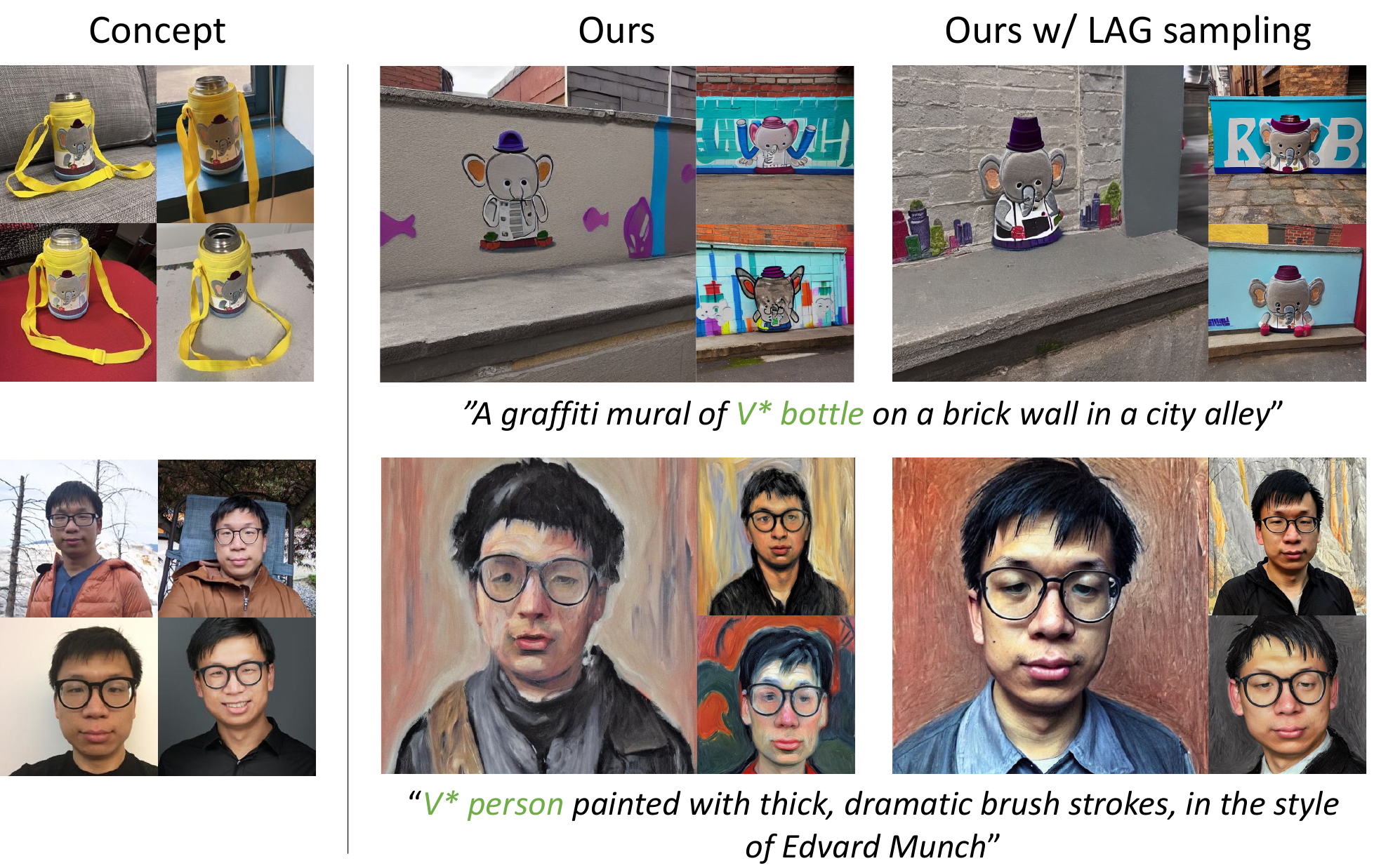}
    \caption{Samples for \textit{artistic style} prompts using personalized residuals with and without LAG sampling where corresponding pairs are generated using the same input noise map.}
    \label{fig:artistic-style}
\end{figure*}

\begin{figure*}
    \centering
    \includegraphics[height=0.42\textheight]{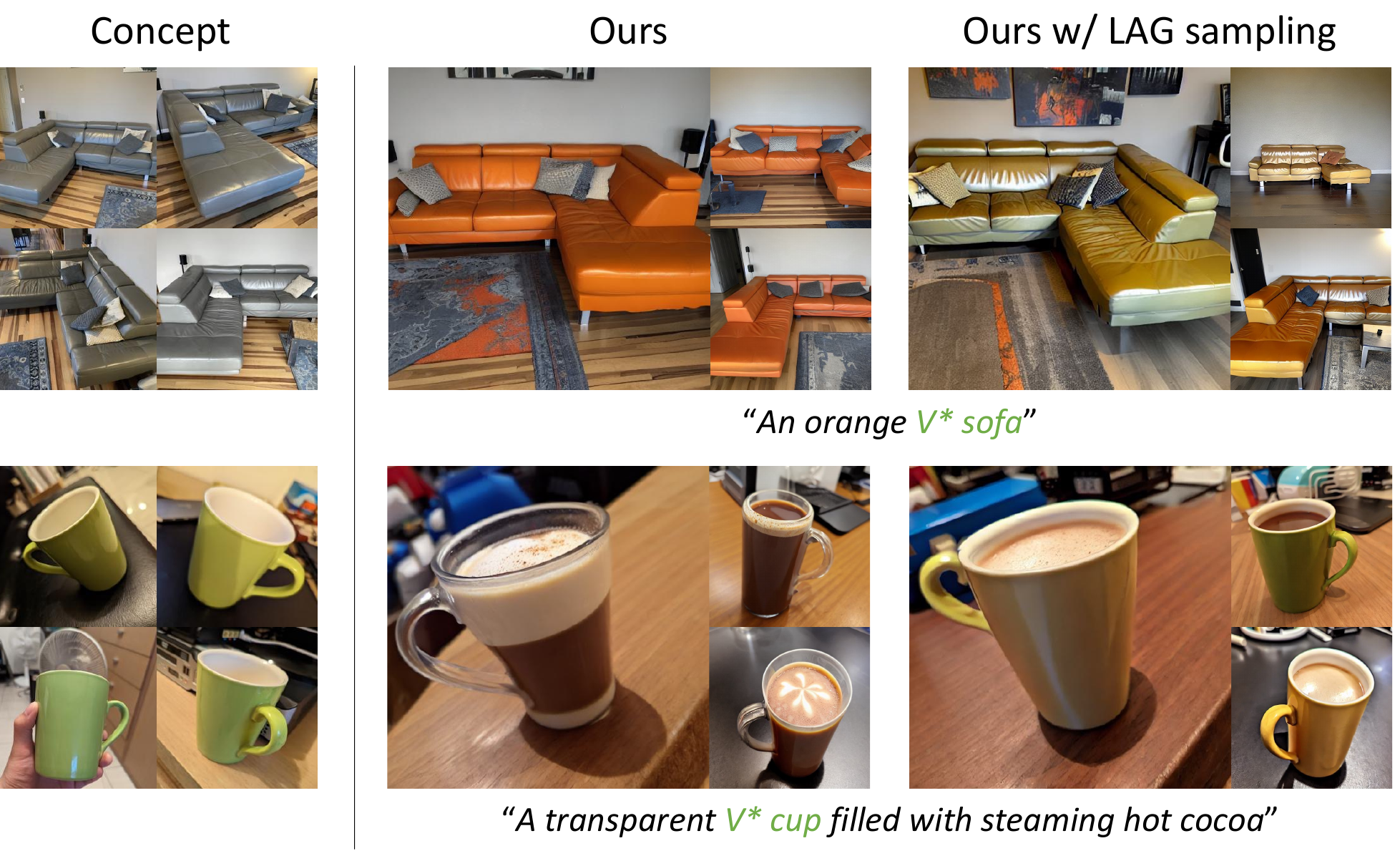}
    \caption{Samples for \textit{change attribute} prompts using personalized residuals with and without LAG sampling where corresponding pairs are generated using the same input noise map.}
    \label{fig:change-attribute}
\end{figure*}

\begin{figure*}
    \centering
    \includegraphics[height=0.42\textheight]{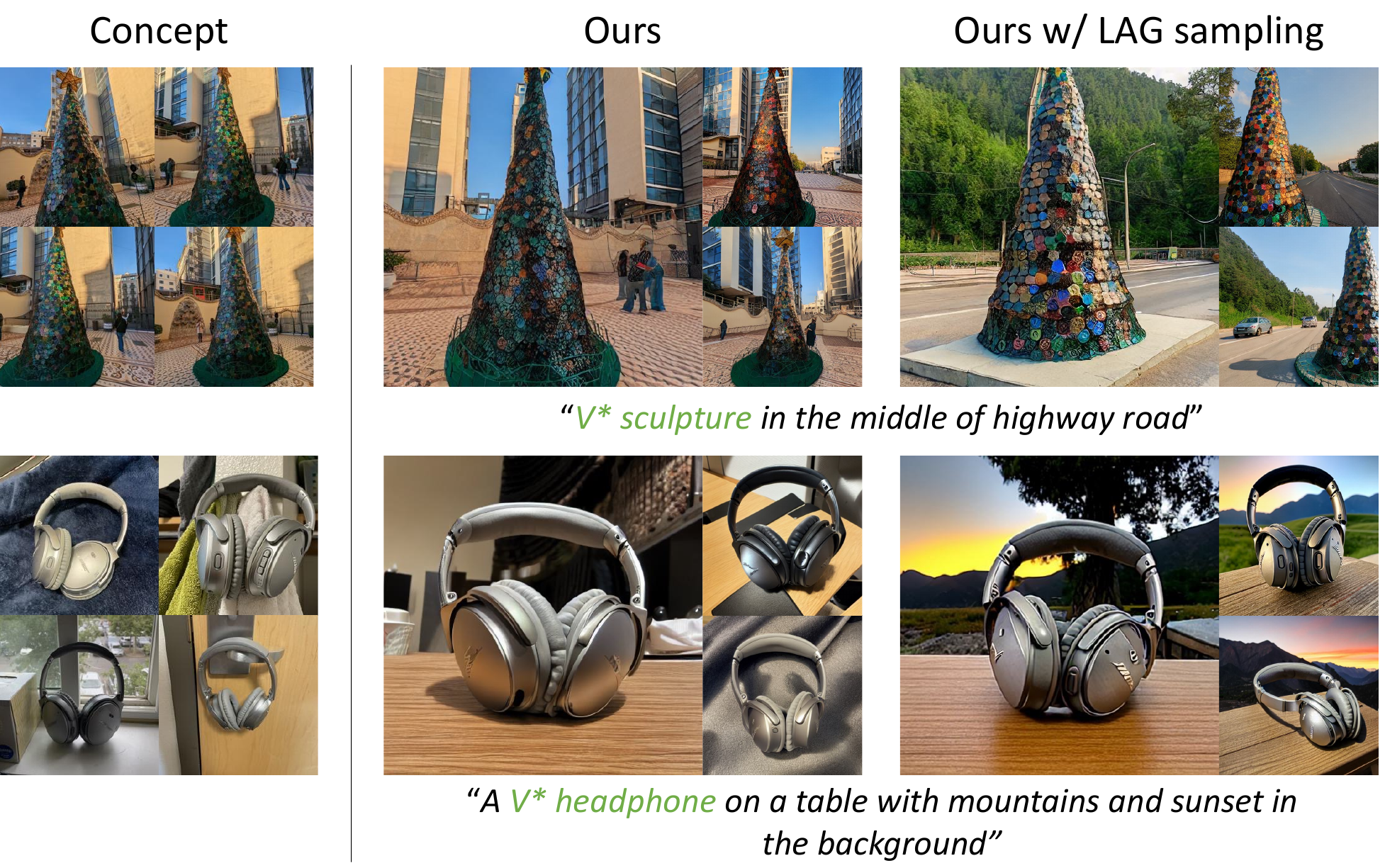}
    \caption{Samples for \textit{change background} prompts using personalized residuals with and without LAG sampling where corresponding pairs are generated using the same input noise map.}
    \label{fig:change-background}
\end{figure*}

\begin{figure*}
    \centering
    \includegraphics[height=0.42\textheight]{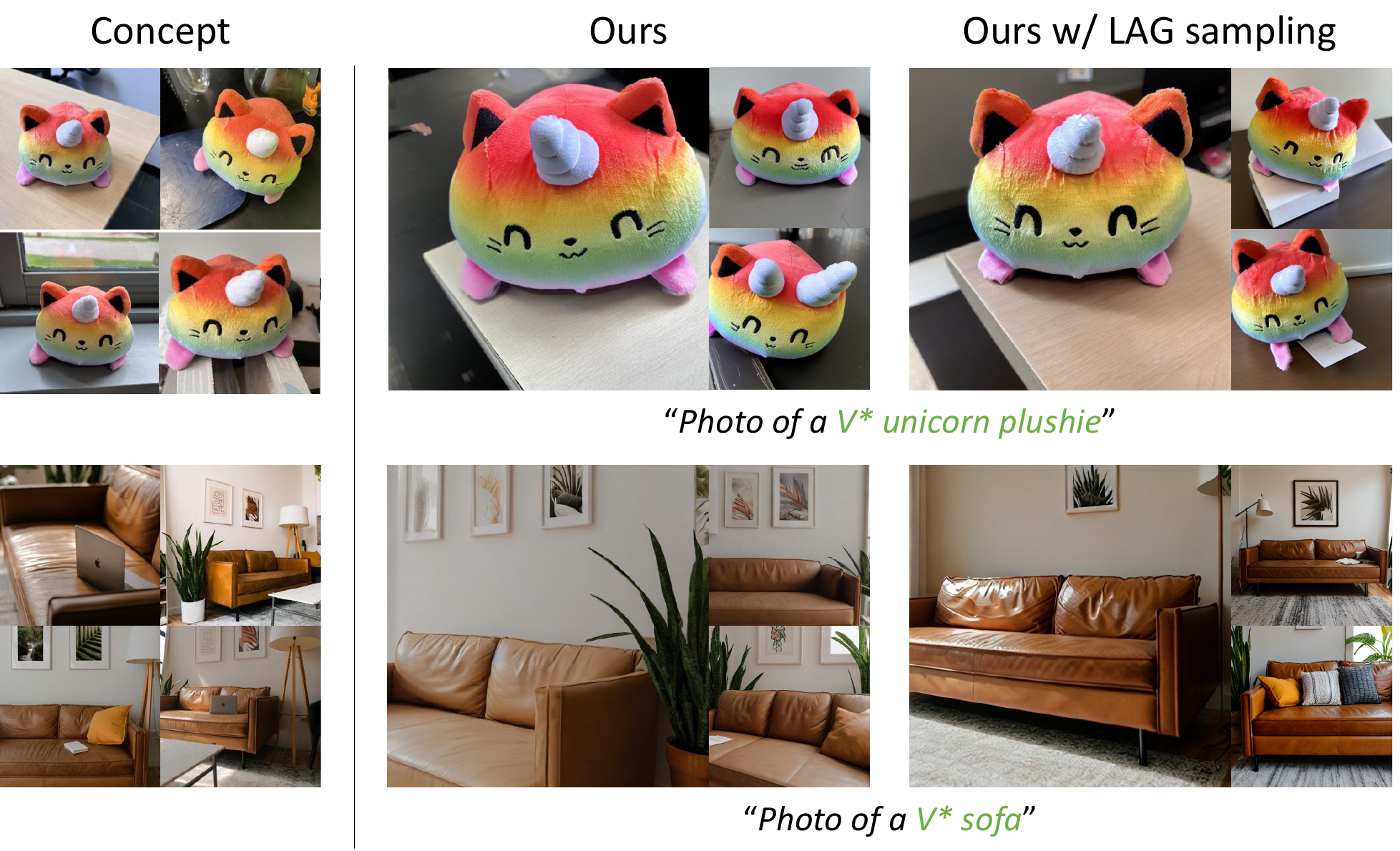}
    \caption{Samples for \textit{identity} prompts using personalized residuals with and without LAG sampling where corresponding pairs are generated using the same input noise map.}
    \label{fig:identity}
\end{figure*}

\begin{figure*}
    \centering
    \includegraphics[height=0.42\textheight]{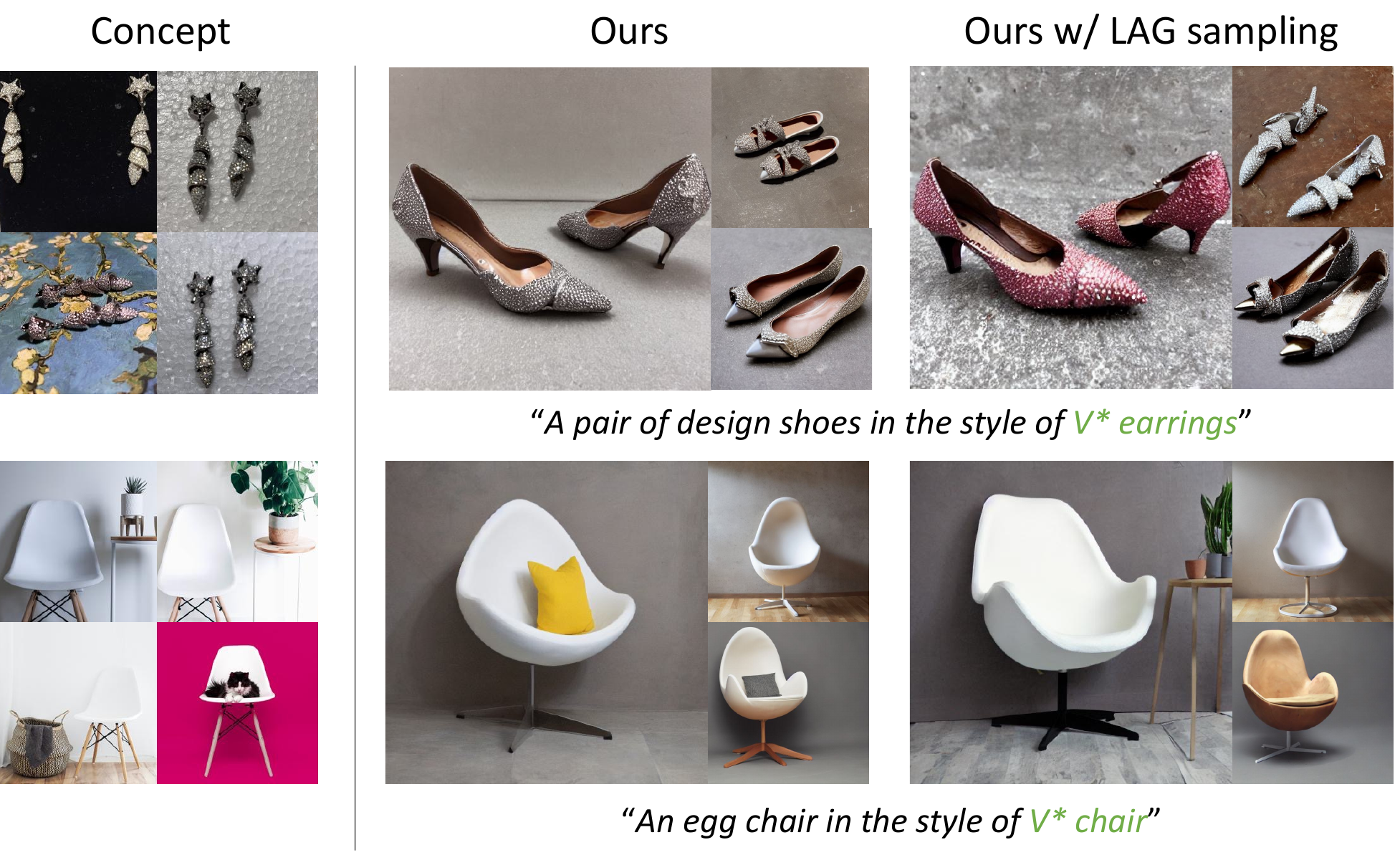}
    \caption{Samples for \textit{object in style of \texttt{V*}} prompts using personalized residuals with and without LAG sampling where corresponding pairs are generated using the same input noise map.}
    \label{fig:object-in-style-of}
\end{figure*}

\begin{figure*}
    \centering
    \includegraphics[height=0.95\textheight]{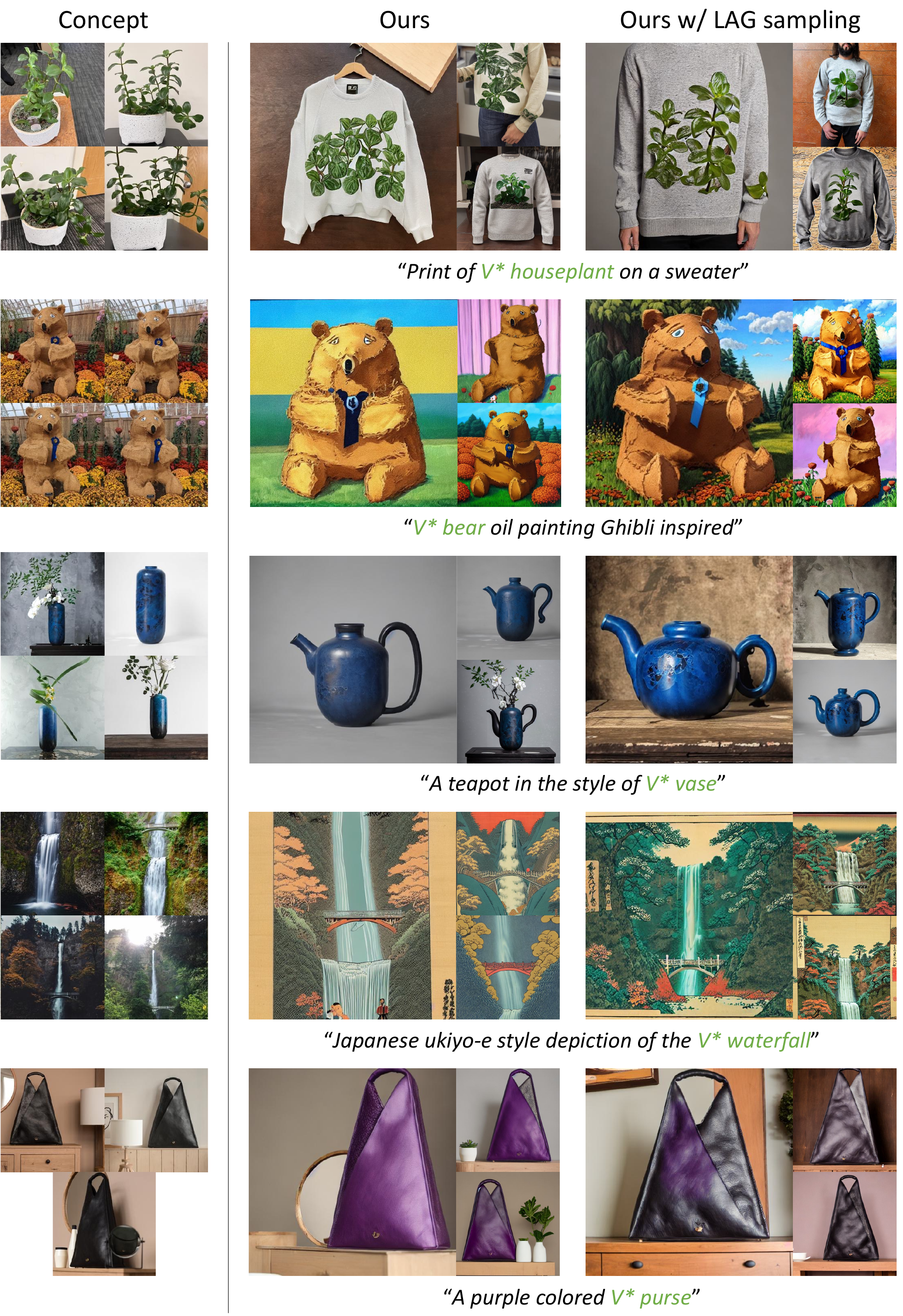}
    \caption{Samples generated using personalized residuals with and without LAG sampling.}
    \label{fig:additional-samples}
\end{figure*}

\begin{figure*}
    \centering
    \includegraphics[height=0.95\textheight]{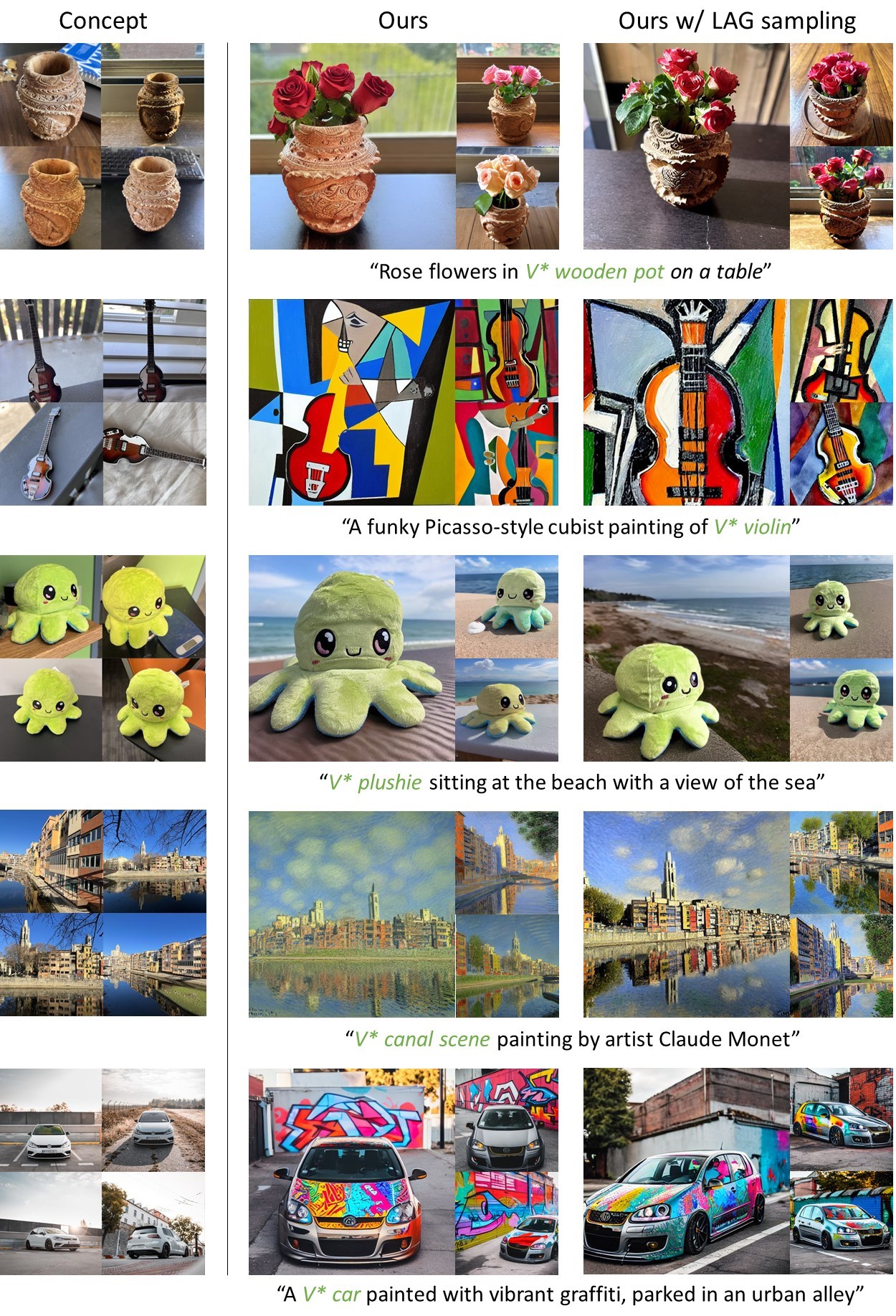}
    \caption{Samples generated using personalized residuals with and without LAG sampling.}
    \label{fig:additional-samples1}
\end{figure*}

In \Cref{fig:add-object,fig:artistic-style,fig:change-attribute,fig:change-background,fig:identity,fig:object-in-style-of} we directly compare examples from each of the six prompt categories using the two sampling methods by generating corresponding pairs using the same starting noise maps. Additional qualitative samples can be found in \Cref{fig:additional-samples,fig:additional-samples1}.

\begin{figure*}
    \centering
    \begin{subfigure}[T]{0.45\textwidth}
        \centering
        \includegraphics[width=\textwidth]{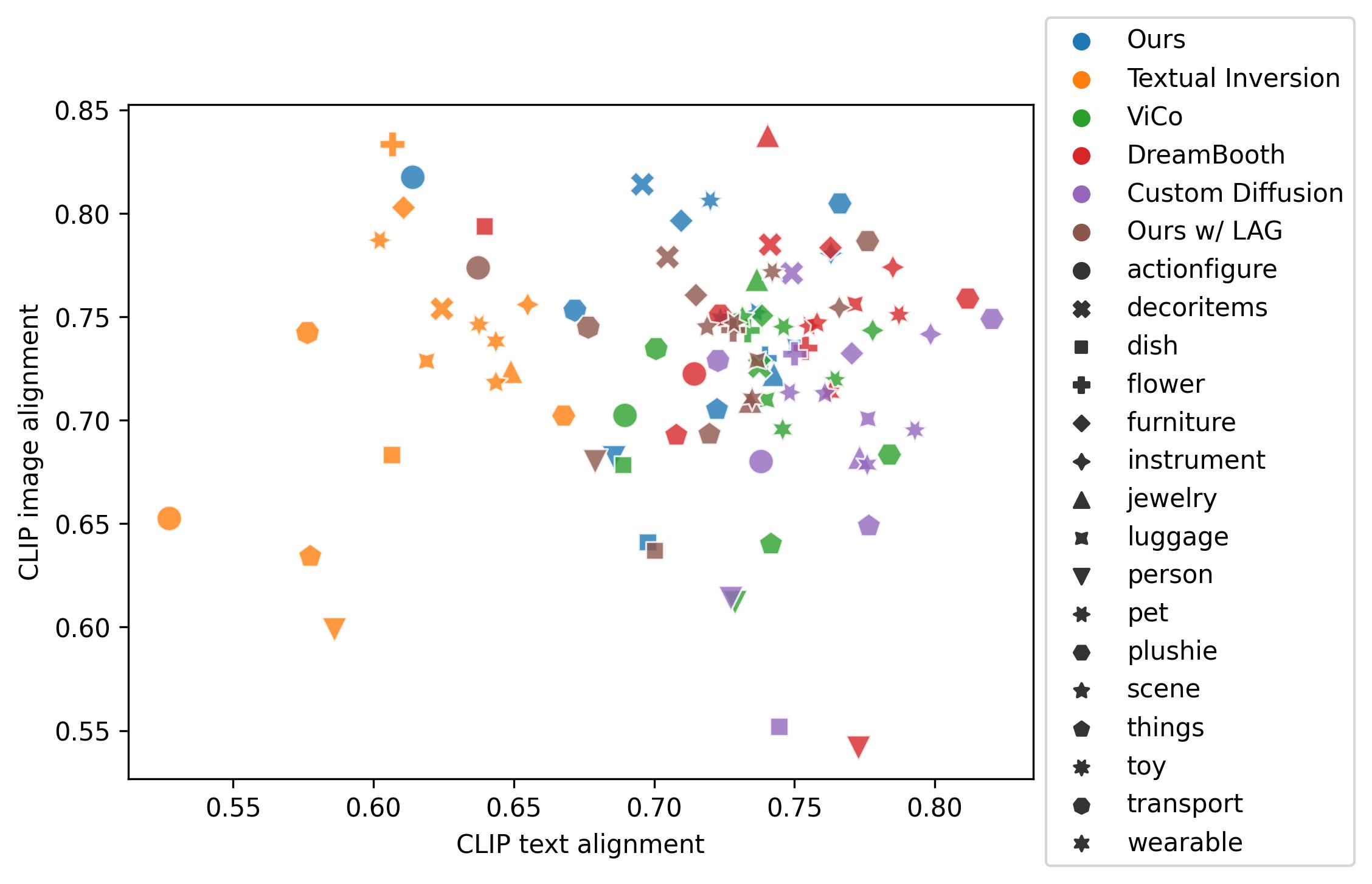}
        \caption{Plot of CLIP image alignment vs. CLIP text alignment.}
        \label{fig:clipimg-vs-cliptxt}
    \end{subfigure}
    \hfill
    \begin{subfigure}[T]{0.45\textwidth}
        \centering
        \includegraphics[width=\textwidth]{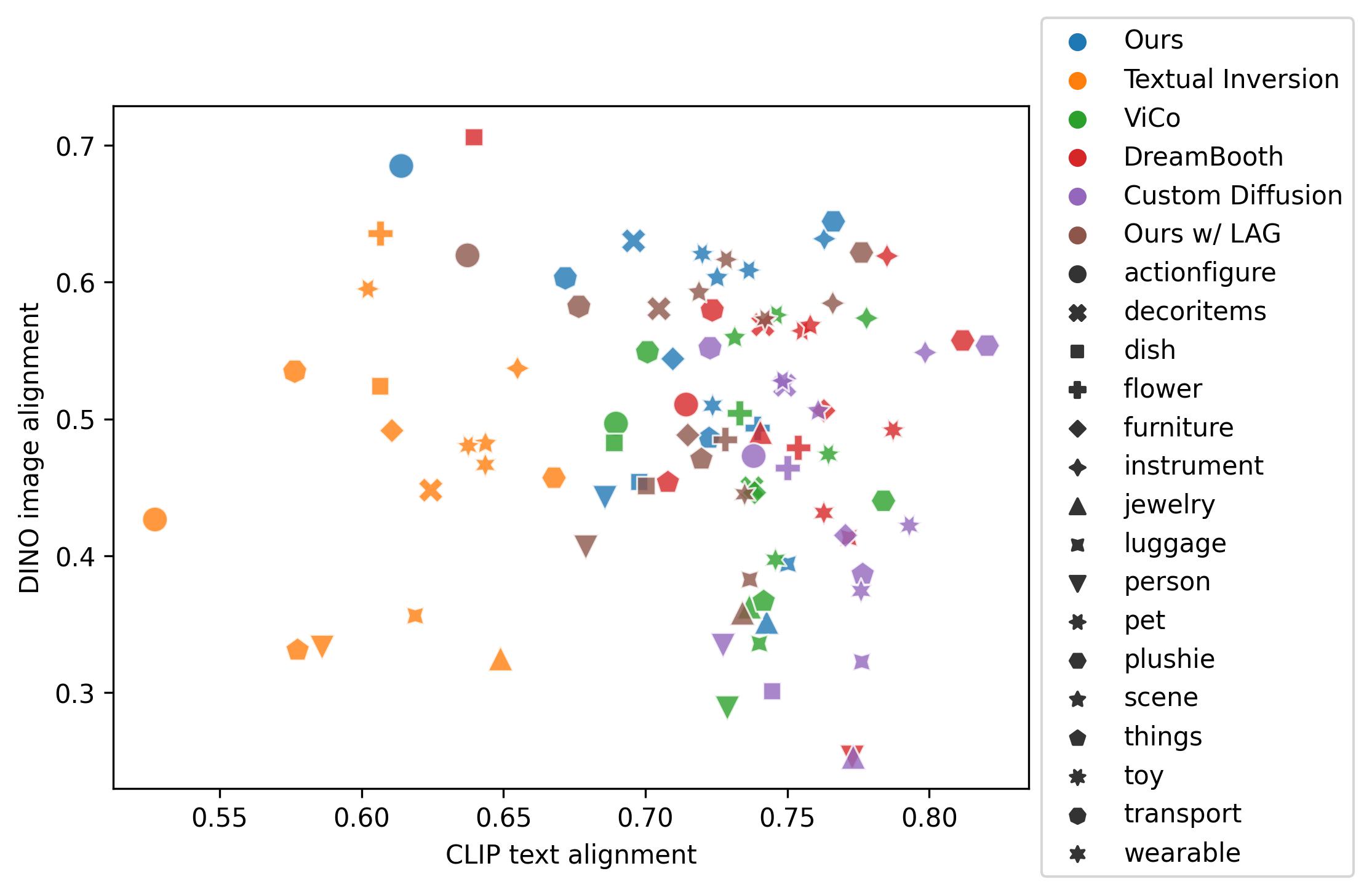}
        \caption{Plot of DINO image alignment vs. CLIP text alignment.}
        \label{fig:dinoimg-vs-cliptxt}
    \end{subfigure}
    \caption{For each method, we plot the either CLIP or DINO image alignment scores against CLIP text alignment scores averaged across the concepts within each of the 16 categories of CustomConcept101.}
    \label{fig:category-scores}
\end{figure*}

We plot CLIP/DINO image alignment scores against CLIP text scores, averaged across concepts within the the 16 categories of CustomConcept101, for each method from \Cref{sec:experiments}.

Additionally, we compare our method to an unofficial implementation\footnote{\url{https://github.com/ChenDarYen/Key-Locked-Rank-One-Editing-for-Text-to-Image-Personalization/}} of Perfusion \cite{tewel2023key} (an official version is not publicly available). We followed the experimental setup and hyperparameter values described by the original authors, but note that we were unable to reproduce the quality of the results shown in the paper: CLIP text 0.6879, CLIP image 0.5669, DINO image 0.2228.

\section{Effect of macro class choice} \label{sec:appendix-macro}

\begin{table}[h]
\caption{We compute the nearest neighbor (NN) in CLIP embedding space for each concept among all WordNet nouns. We compare our method using different combinations of macro classes during training and sampling.}
\label{tab:macro-comparison}
\begin{adjustbox}{width=\linewidth}
\begin{tabular}{llccc}
\multicolumn{2}{c}{\bf Macro class choice}  &  \multirow{2}{*}{\bf CLIP text}  &  \multirow{2}{*}{\bf CLIP image}  &  \multirow{2}{*}{\bf DINO image} \\ 
\textbf{Training}  &  \textbf{Sampling}  &  &  & \\
\hline
\multirow{2}{*}{CustomConcept101}  &  CustomConcept101  &  0.7193  &  0.7594  &  0.5671 \\
  &  WordNet NN  &  0.7155  &  0.7594  &  0.5671 \\
\midrule
\multirow{2}{*}{WordNet NN}  &  CustomConcept101  &  0.6626  &  0.7798  &  0.5904 \\
  &  WordNet NN  &  0.6869  &  0.7798  &  0.5904
\end{tabular}
\end{adjustbox}\vspace{-1em}
\end{table}

For each concept in CustomConcept101, we compute the mean CLIP image embedding of its reference images and calculate the cosine similarity against the CLIP text embedding for each of the 117k nouns within WordNet. We train our method and/or sample using the WordNet noun with the highest similarity and compare with using the provided macro class from CustomConcept101 during training and/or sampling in \Cref{tab:macro-comparison}. We observe that using the WordNet nearest neighbor as the macro class leads to higher image alignment and lower text alignment compared to the CustomConcept101-provided macro class.

Selecting the ``best'' macro class for concepts can be challenging and given that it can lead to noticeable changes in alignment metrics, an automatic heuristic for choosing a suitable macro class would be helpful to users. We leave the designing of such a heuristic as future work.

\section{Ablation study: rank value} \label{sec:appendix-rank}

\begin{table}[h]
\caption{Quantitative evaluations for varying the rank of the learned residuals. $m_i$ is the dimension of the weight of the projection layer in transformer block $i$.}
\label{tab:ablation-rank}
\begin{adjustbox}{width=\linewidth}
\begin{tabular}{lccc}
\multicolumn{1}{l}{\bf Rank}  &  \multicolumn{1}{c}{\bf CLIP text}  &  \multicolumn{1}{c}{\bf CLIP image}  &  \multicolumn{1}{c}{\bf DINO image} \\ 
\hline
1  &  0.7398  &  0.6809  &  0.4148 \\
8  &  0.7054  &  0.7402  &  0.5239 \\
16  &  0.6926  &  0.7573  &  0.5513 \\
32  &  0.6832  &  0.7701  &  0.5713 \\
64  &  0.6704  &  0.7798  &  0.5865 \\
128  &  0.6544  &  0.7938  &  0.6053 \\
$0.025m_i$  &  0.6889  &  0.7622  &  0.5595 \\
\hdashline
Ours ($0.05m_i$)  &  0.7193  &  0.7594  &  0.5671
\end{tabular}
\end{adjustbox}\vspace{-0.5em}
\end{table}

We evaluate different values for the rank of the learned residuals in \Cref{tab:ablation-rank} and observe that text alignment is inversely proportional to the rank and image alignment is directly proportional. Since the dimensions of the conv weight matrix varies across the transformer blocks within the U-Net, we believe that calculating the rank with respect to the dimensions is the better approach over setting a fixed value across all layers, which is empirically validated by the results with our proposed formula achieving a better balance of image and text alignment.

\end{document}